\documentclass{article} 
\usepackage{collas2024_conference,times}
\usepackage{easyReview}

\usepackage{graphicx}
\usepackage{booktabs}
\usepackage{multirow}
\usepackage{xcolor}
\usepackage{colortbl}
\usepackage{amsmath}

\usepackage{amssymb}
\newcommand\mdoubleplus{\mathbin{+\mkern-10mu+}}

\usepackage{amsmath,amsfonts,bm}

\def\eqref#1{equation~\ref{#1}}

\def\1{\bm{1}}

\DeclareMathAlphabet{\mathsfit}{\encodingdefault}{\sfdefault}{m}{sl}
\SetMathAlphabet{\mathsfit}{bold}{\encodingdefault}{\sfdefault}{bx}{n}

\usepackage{hyperref}
\hypersetup{
    colorlinks=true,
    linkcolor=red,
    filecolor=magenta,
    urlcolor=blue,
    citecolor=purple,
    pdftitle={Overleaf Example},
    pdfpagemode=FullScreen,
    }

\title{Label Augmentation  for Neural Networks Robustness}

\author{Fatemeh Amerehi \\
University of Limerick, Ireland \\
\texttt{fatemeh.amerehi@ul.ie} \\
\And 
Patrick Healy   \\
University of Limerick, Ireland \\
\texttt{patrick.healy@ul.ie} \\
}

\collasfinalcopy 

\begin{document}
\maketitle

\begin{abstract}
Out-of-distribution generalization can be categorized into two types: common perturbations arising from natural variations in the real world and adversarial perturbations that are intentionally crafted to deceive neural networks. While deep neural networks excel in accuracy under the assumption of identical distributions between training and test data, they often encounter out-of-distribution scenarios resulting in a significant decline in accuracy. Data augmentation methods can effectively enhance robustness against common corruptions, but they typically fall short in improving robustness against adversarial perturbations. In this study, we develop Label Augmentation (LA), which enhances robustness against both common and intentional perturbations and improves uncertainty estimation. Our findings indicate a Clean error rate improvement of up to 23.29\% when employing LA in comparisons to the baseline.
Additionally, it enhances robustness under common corruptions benchmark by up to 24.23\%.
When tested against FGSM and PGD attacks,
improvements in adversarial robustness are noticeable, with enhancements of up to 53.18\% for FGSM and 24.46\% for PGD attacks.

\end{abstract}

\section{Introduction}

\begin{figure}[h]
  \centering

    \includegraphics[width=0.45\linewidth]{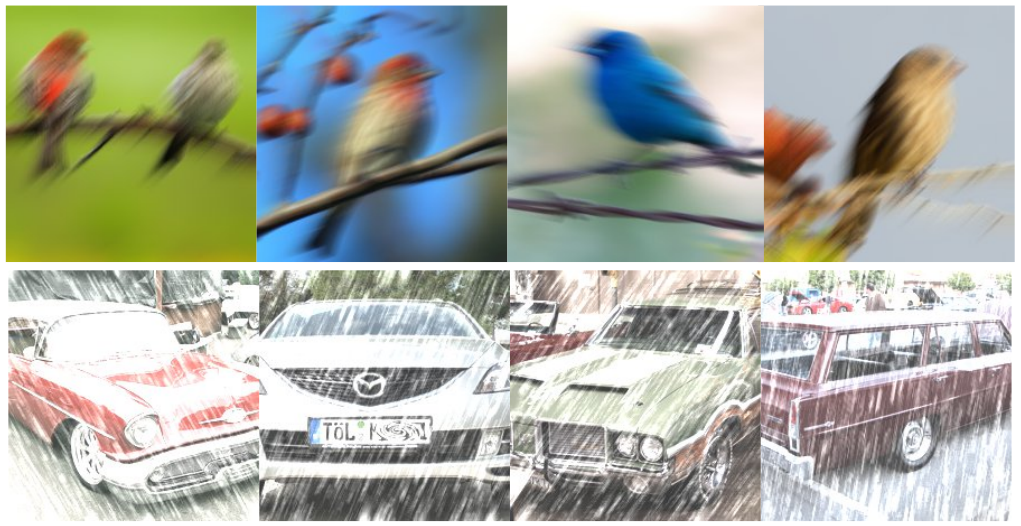} 
   \caption{What do you see when looking at the images?}
   \label{fig:blur_snow}
\end{figure}

Real-world objects exhibit a diverse array of intertwined attributes. While certain characteristics, such as the class identity of the object, are permanent, others, like the lighting conditions or pose of the object, are transient~\citep{gabbay2019demystifying}. In fact, how would you interpret the images in Fig.~\ref{fig:blur_snow}? If you were to present them to someone and ask about their interpretation, they would likely identify~\textbf{\textit{blurry}} birds and cars in~\textbf{\textit{snow}}. Despite the variations in the images, we are capable of effectively distinguishing between the class identity and the transient attributes of an object.

We already assign names and labels for objects around us, but we also have names for concepts like brightness, warmth, noisiness, and many more.
The birds or cars themselves are unchanged, but the sharpness and colors are different. Essentially, the name/labels remain invariant to us, but we still recognize that there exist some other elements that differ from each other. The process of training a machine to make similar distinctions among various attributes in observed data is referred to  as disentanglement. 
It aims to find latent representations that adeptly separate the explanatory factors contributing to variations in the input data~\citep{bengio2013representation}.

Disentangled representations have been shown to improve generalization to unseen scenarios in both generative and discriminative tasks~\citep{gabbay2019demystifying, eom2019learning,trauble2021disentangled}. Deep Neural Networks (DNNs) generalize well under the assumption of Independent and Identically Distributed (IID) data, where both training and test datasets come from the same distribution. Yet, high IID accuracy does not guarantee out-of-distribution (OOD) generalization where train and test distributions mismatch ~\citep{ liu2021towards}. 

Common corruptions~\citep{hendrycks2019benchmarking} and adversarial perturbations~\citep{goodfellow2014explaining} are two examples of OOD scenarios leading to performance deterioration.  A widely used approach to mitigate performance drop is to incorporate data augmentation into the training pipeline~\citep{shorten2019survey}. While data augmentation enhances model robustness, current methods tend to improve either common corruption or adversarial perturbation individually, rather than concurrently enhancing both. 

Beyond vulnerability to distributional shifts, another common issue is miscalibration—the tendency of models to generate overconfident predictions when the training examples are IID. This overconfidence is further intensified under OOD scenarios~\citep{ovadia2019can}. In this study, for enhanced robustness, we present a simple yet effective method using Label Augmentation (LA) for disentangling the class of an object from irrelevant noise. The LA proves effective in enhancing calibration and robustness against both common and intentional perturbations of input data.

\section{Related Works}
\label{literature}

Focusing on vision models, we review relevant literature on augmentation methods for robustness against distributional shifts, including adversarial attacks, and common corruptions, alongside calibration.

\textbf{Augmentation methods for robustness under distribution shift.} Vision models often experience a drop in performance under common or intentional perturbations of images~\citep{hendrycks2019benchmarking, szegedy2013intriguing}. For instance, they show vulnerability to blur and Gaussian noise~\citep{vasiljevic2016examining, dodge2016understanding}, as well as factors such as brightness and contrast~\citep{hendrycks2019benchmarking}, occlusion~\citep{zhong2020random}, and small translations or rescalings of the input data~\citep{azulay2018deep}. Additionally, when the model encounters adversarial perturbations, its performance tends to suffer even more~\citep{goodfellow2014explaining, papernot2016transferability, tramer2017space, athalye2018obfuscated}.

To mitigate performance degradation caused by common corruptions, a commonly employed strategy is the incorporation of label-preserving image augmentation into the training pipeline~\citep{shorten2019survey}. In the simplest form, data augmentations translate to  simple transformations such as horizontal flipping, color shift, and random cropping~\citep{krizhevsky2012imagenet, he2016deep}. 
A more complex array of augmentations includes techniques such as random erasing~\citep{devries2017improved, zhong2020random}, neural style transfer~\citep{jackson2019style, geirhos2018imagenet}, image mixing~\citep{zhang2017mixup,inoue2018data, summers2019improved, hong2021stylemix, yao2022improving}, training with noise~\citep{lopes2019improving,rusak2020simple}, randomized manipulations of images~\citep{xu2023comprehensive}, combination and mixing of augmentation chains~\citep{hendrycks2019augmix,modas2022prime},
or search for an optimal augmentation policy~\citep{cubuk2019autoaugment}.

%

Defense mechanisms to tackle adversarial examples—a carefully crafted perturbations to mislead a classifier— include defensive distillation \citep{papernot2016distillation}, feature squeezing \citep{xu2017feature}, adversarial detection \citep{metzen2017detecting,pang2018towards,deng2021libre}, gradient regularization \citep{tramer2017ensemble,wu2020adversarial}, and adversarial training \citep{goodfellow2014explaining,madry2017towards}. Among these, the most effective strategy is adversarial training, which involves augmenting training data with adversarial examples to enhance its robustness against attacks or to reduce its test error on clean inputs~\citep{goodfellow2014explaining, kurakin2016adversarial, moosavi2016deepfool, ford2019adversarial, bai2021recent}. 

Adversarial examples could be augmented in various ways, including incorporating synthetic data \citep{gowal2021improving,wang2023better}, unlabeled data \citep{carmon2019unlabeled,deng2021improving}, injecting noise to the hidden layers~\citep{qin2022understanding}, adversarial mixture of transformations \citep{wang2021augmax}, or reconfiguration of the low and high-frequency components of intermediate feature representations \citep{bu2023towards}. Other methods introduce weight perturbation to enhance model robustness~\citep{wu2020adversarial}, regulating gradient growth to prevent robust overfitting during multi-step adversarial training~\citep{li2022subspace}, or ensemble training to mitigate vulnerabilities across sub-models while preserving comparable accuracy on clean data~\citep{cai2023ensemble}.

The effectiveness of adversarial training depends on the choice of adversarial examples. For instance, training exclusively with Fast Gradient Sign Method (FGSM)~\citep{goodfellow2014explaining} enhances robustness against non-iterative attacks but lacks robustness against iterative attacks such as Projected Gradient Descent (PGD) attack~\citep{kurakin2016adversarial, madry2017towards}. Whether adversarial training enhances robustness against common corruptions has conflicting views in the literature. While some studies suggest a positive correlation~\citep{ford2019adversarial, kireev2022effectiveness}, others argue that adversarial robustness and robustness to common perturbations are independent~\citep{laugros2019adversarial}.


\textbf{Calibration.} Despite performing well in generalization and prediction under the IID setting, DNNs  often produce overconfident results, which worsen even more in OOD settings~\citep{guo2017calibration, ovadia2019can, gawlikowski2021survey}. Well-calibrated uncertainty estimates indicate when the output of models is reliable and when it is questionable. Temperature scaling with a validation set~\citep{guo2017calibration}, or ensembling predictions from independently trained classifiers on the entire dataset with random initialization~\citep{lakshminarayanan2017simple}, improves calibration. Using soft labels~\citep{szegedy2016rethinking}—a weighted average of one-hot training labels and a uniform distribution over targets—often prevents the network from becoming overly confident in specific labels, thus reducing calibration errors~\citep{muller2019does,lukasik2020does}.

\begin{figure}[]
\begin{center}

\includegraphics[scale=0.42]{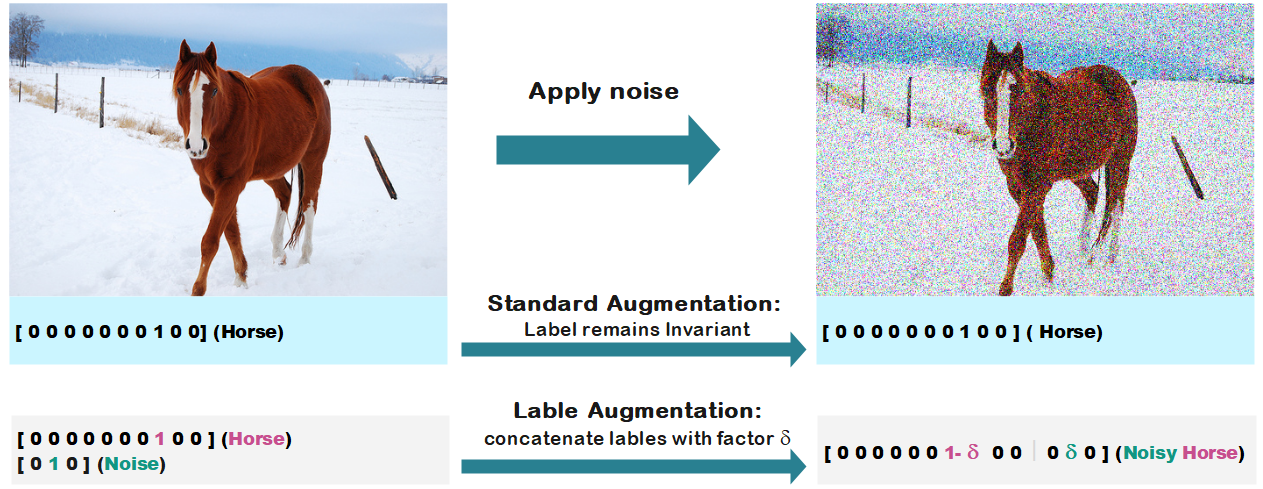}
\label{fig:LA}
\end{center}
\caption{The Cifar10 dataset includes 10 classes representing airplanes, cars, birds, cats, deer, dogs, frogs, horses, ships, and trucks. The one-hot label for horses is [0 0 0 0 0 0 0 \textcolor{red}{1} 0 0]. Considering three distinct augmentation operation classes like contrast, noise, and blur; the one-hot label for noise is [0 \textcolor{teal}{1} 0]. In standard augmentation, labels remain invariant. When applying Label Augmentation with a smoothing factor $\delta$, the resulting label for noisy image of a horse is [0 0 0 0 0 0 0 \textcolor{red}{$1- \delta$} 0 0 0 \textcolor{teal}{$\delta$} 0]. This maintains invariance with original categories while distinguishing between more abstract concepts, such as noisy and noise-free inputs.}
\end{figure} 

\section{Label Augmentation}
\label{sec:Label Augmentation}

We now introduce our central idea.  

Let $\mathcal{O} = \{o_i\}_{i=1}^{M}$ represent a set of $m$ label-preserving augmentation operations, each of which, when applied to an input data, introduces certain effects to it. Let $\mathcal{Z} = \{z_i\}_{i=1}^{M}$ be the one-hot encoded name for each of the operations.
Given a collection of objects $\mathcal{X} = \{x_i\}_{i=1}^N$ and a set of labels $\mathcal{Y} = \{y_i\}_{i=1}^K $, we humans assign a label $y_i$ to each object $x_i$ based on the attributes we observe in them.  Further, let $\mathcal{O}(\mathcal{X})$ represent operations within $\mathcal{O}$ that are applied to objects in $\mathcal{X}$.

If we select an operation $o_j \in  \mathcal{O}$ and apply it to $x_i \in \mathcal{X}$ as the certain attributes of the $x_i$ are affected—despite  the identity of the class object remains the same for each elements in \(\mathcal{O}(\mathcal{X})\)—we no longer assign the same label \( y_i \) to the transformed objects/images (revisit Fig.~\ref{fig:blur_snow}). Instead, we assign a richer name that incorporates both the class identity and the effect. In essence, we disentangle class identity from transformations/operations. 

In a task of \(K\)-class classification, the goal is to model the mapping from the input data \(x_i\) to its corresponding class label \(y_i\) through a DNN \(f : \mathcal{X} \to \mathcal{Y}\).  
Typically, this involves using a softmax output layer and cross-entropy loss 
to quantify  the dissimilarity between \(y_i\), the true class distribution (one-hot encoded), and \(p_i\), the softmax of predictions. The cross-entropy loss,  is defined as \(\mathcal{L}_{CE}(y_i, p_i) = -\sum_{k=1}^K y_{ik} \log p_{ik}\), where \(y_{ik}\) represents the \(k\)-th element of the true class distribution \(y_i\), and \(p_{ik}\) denotes the \(k\)-th element of the predicted class distribution \(p_i\).

To extend the model generalization capability to OOD data, existing augmentation methods train with augmented input  \(\mathcal{O}(\mathcal{X})\) while assigning the same label to transformed and untransformed input data to help the model learn representations that remain invariant to a set of data augmentations. Essentially, existing techniques aim to find a mapping \(f : \mathcal{X} \cup \mathcal{O}(\mathcal{X}) \to \mathcal{Y}\).  Given the distinction we make in our naming between \( \mathcal{X} \) and \( \mathcal{O}(\mathcal{X}) \), and considering the use of such augmented inputs in training DNNs, would it not be advantageous  to explicitly communicate to the model that labels differ in additional factors beyond class identity? To enable this, we employ Label Augmentation (LA).

In LA, the objective is to maintain invariance to the input class category \(y_i\), while simultaneously enabling the distinctions between \(x_i\) and its various transformed versions. To achieve this, after any transformation \(o_j\) on input data $x_i$, we simply concatenate the two one-hot labels \(y_i\) and \(z_j\) to each other with a factor of \(\delta\).  In other words, whenever we augment the input data, we augment labels as well. Specifically, the labels assigned to $\Tilde{x_i} = o_j(x_i)$ are defined as in Eq.~\ref{eq:LA_label}, which represent a vector of length $K+M$. The label $\Tilde{y_i}$ has the value of $1-\delta$ at position $i$ and the value of $\delta$ at position $K+j$. In other words, LA aims to find a more comprehensive mapping \(f : \mathcal{X} \cup \mathcal{O}(\mathcal{X}) \to \text{Concat}_{\delta}[\mathcal{Y} , \mathcal{Z}] \) that maps augmented collections of input to the augmented labels. This is shown in Fig.~\ref{fig:LA}.

\begin{equation}\label{eq:LA_label}
\Tilde{y_i} = \text{Concat}[(1-\delta)y_i, \delta z_j]
\end{equation}

We consider identity transformation with $\delta=0$ as a specific case that represents untransformed input data. In case of transformation, we select \(\delta\) to be a small value in order to prevent excessive deviation of the model towards the augmented label. The value of \(\delta\) is drawn from a uniform distribution: \(\delta \sim \text{U}(0.05, 0.1)\). 

To ensure the same dimensionality and maintain the class identity $y_i$ for any untransformed input data $x_i$, we simply expand the one-hot labels $y_i$ from $K$ dimensions to $K+M$. At position $i$, we assign 1, as before, to represent the class identity. In LA training, the loss is computed as $\mathcal{L}_{LA}(\Tilde{y}_i, \Tilde{p_i}) = -\sum_{k=1}^{K+M} \Tilde{y}_{ik} \log \Tilde{p}_{ik}$, where $\Tilde{p}_i$ denotes the softmax of predictions for $\Tilde{x_i}$.

In the following section, we show that the act of assigning names to operations and augmenting labels leads to better generalization compared to traditional augmentation. Furthermore, as we will demonstrate, this helps achieve better robustness to both common and intentional perturbations, as well as improved calibration.

\begin{figure}[h]
\begin{center}
\includegraphics[scale=0.48]{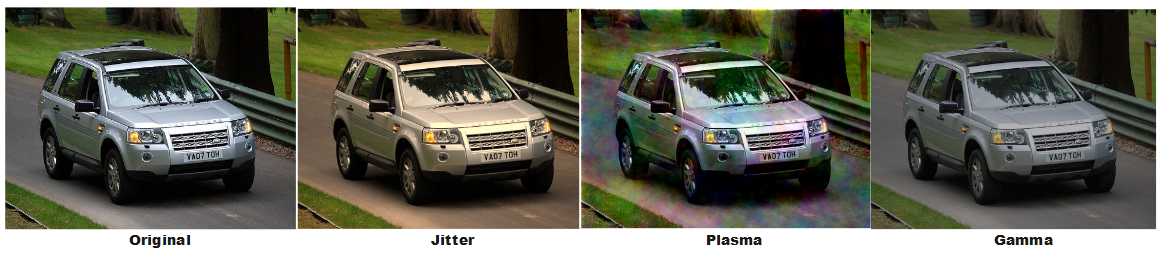}
\label{fig:sample}
\end{center}
\caption{Examples of augmentation operations applied in Label Augmentation.}
\end{figure}

\section{Experimental Setup}
\label{exp}

In the following, we elaborate on the dataset, training configuration, the networks employed, and the evaluation metrics for assessing both robustness and calibration. Afterwards, we present the results and analysis.

\subsection{configurations and metrics}

\textbf{Datasets.}
We utilize the CIFAR-10 and CIFAR-100~\citep{krizhevsky2009learning} datasets. Both datasets contain 50,000 training images and 10,000 testing images of size 32 × 32 × 3.
To assess the robustness of models against common data shift, we evaluate on CIFAR-10-C and CIFAR-100-C
benchmark~\citep{hendrycks2019benchmarking}. These datasets are created by introducing various distortions to the original CIFAR-10 and CIFAR-100 datasets, and contains a total of 15 corruptions of types such as noise, blur, weather, and digital distortions. Each distortion is incorporated at severity levels \(1 \leq s \leq 5\). In the following, we refer to these datasets as CIFAR and CIFAR-C, respectively.

\textbf{Robustness metrics.} In Tables~\ref{tab:cf10_LC}, ~\ref{tab:cf10_No_LC}, and~\ref{tab:allCF10}, the Clean Error represents the standard classification error on uncorrupted test data. For a given corruption \(c\), the error rate at corruption severity \(s\) is denoted as \(\text{E}_{c,s}\). Taking the average error across these severities \(s\); the corruption error \(\text{CE}_c\), is computed as $\text{CE}_c = \frac{1}{5} \sum_{s=1}^{5} \text{E}_{c,s}$. Finally,  the mean Corruption Error $\text{mCE} = \frac{1}{15} \sum_{c=1}^{15} \text{CE}_{c}$ is the average of all 15 corruption errors. This gives one value for robustness comparisons against common corruptions~\citep{hendrycks2019benchmarking,hendrycks2019augmix}.

To measure robustness against adversarial perturbations, we employ FGSM~\citep{goodfellow2014explaining} and 40-step iterative PGD~\citep{ madry2017towards} attacks, both with $L_\infty$ constraints using two budgets $\varepsilon = 0.03$ and $\varepsilon = 0.3$. We utilize the implementations provided by the \textit{cleverhans} 4.0 library~\citep{papernot2018cleverhans}. 

\textbf{Calibration metric.} A classifier is considered calibrated when it can consistently predict their accuracy~\citep{guo2017calibration}. For instance, with 100 predictions, each at a confidence level of 0.7, we expect 70 correct classifications. We evaluate the calibration of the network using the Expected Calibration Error (ECE)~\citep{guo2017calibration} and Root Mean Square (RMS) Calibration Error. Given the finite size of the test sets, ECE and RMS Calibration Error are estimated by grouping all $n$ test examples into $M$ equal size bins, ordered by prediction confidence—the winning softmax score. 

Let \( B_m \) represent the set of samples whose predictions fall into bin \( m \). The accuracy and confidence of \( B_m \) is defined as $\text{acc}(B_m) = \frac{1}{|B_m|} \sum_{i \in B_m} 1(\hat{y}_i = y_i)$ and $\text{conf}(B_m) = \frac{1}{|B_m|} \sum_{i \in B_m} \hat{p}_i$, respectively. Here, $\hat{y_i}$ and $y_i$ represent the predicted and ground-truth labels for input data $x_i$, and $\hat{p}_i $ is the confidence—winning score—of sample \( i \). The ECE and RMS errors is then defined as in Eq.~\ref{eq:ECE} and Eq.~\ref{eq:RMS}, respectively.  We use the implementations provided by the \textit{TorchMetric} 1.4.0 library~\citep{torchmetrics}.

\begin{equation}\label{eq:ECE}
\text{ECE}  = \sum_{m=1}^{M} \frac{|B_m|}{n} \Bigl| \text{acc}(B_m) - \text{conf}(B_m) \Bigr|
\end{equation}
\begin{equation}\label{eq:RMS}
\text{RMS} = \sqrt{\sum_{m=1}^{M} \frac{|B_m|}{n} \Bigl( \text{acc}(B_m) - \text{conf}(B_m) \Bigr)^2}
\end{equation}

\textbf{Training Configuration and hyper-parameter setting.} We run all experiments on a GeForce RTX-3080Ti GPU  with CUDA Version 12.0 using the PyTorch version 2.0.1. To assess robustness across different architectures, we use a standard LeNet~\citep{lecun1998gradient}, a ResNet-50~\citep{he2016deep}, a 40-2 Wide ResNet~\citep{zagoruyko2016wide}, a 32$\times$4d ResNeXt-50~\citep{xie2017aggregated}, and a  Swin Transformer~\citep{liu2021swin}. All networks start with a learning rate of 0.1, which decays by a factor of 0.0001 according to a cosine annealing learning rate~\citep{loshchilov2016sgdr}. Before any augmentations, we preprocess input images through random horizontal flip and cropping. 
In all experiments, we train for 25 epochs with default weights and optimize with stochastic gradient descent with a momentum of 0.9. For both training and evaluation, we set the batch size to 1024. Each experiment in Tables~\ref{tab:allCF10},~\ref{tab:allCF100}, and~\ref{tab:all_MTL_LS}  are conducted three times, and the averages along with their corresponding standard deviations are reported.

\textbf{Baseline comparisons.} We begin by comparing LA with traditional augmentations, selecting a set of label-preserving augmentations such as Plasma noise~\citep{nicolaou2022tormentor}, Planckian Jitter~\citep{zini2022planckian}, and Gamma adjustment, as illustrated in Fig.~\ref{fig:sample}. First, we train the models with these augmentations using LA, and then repeat the training with normal augmentations, without LA. The results of these experiments are presented in Tables~\ref{tab:cf10_LC} and~\ref{tab:cf10_No_LC}, and we will analyze them in the next section.

To measure the effectiveness of LA, we make a comparison between various augmentation techniques, including Mixup~\citep{zhang2017mixup}, Augmix~\citep{hendrycks2019augmix}, and AutoAugment~\citep{cubuk2019autoaugment}. Additionally, we include FGSM~\citep{goodfellow2014explaining} and 10-step iterative PGD~\citep{madry2017towards}, both with $L_\infty$ constraints and $\varepsilon = 0.3$, for comparisons against adversarial training.
Except for LA, in all these experiments, we adjust the last classification layer of networks to output 10 and 100 class categories corresponding to CIFAR-10, and CIFAR-100 datasets, respectively. For LA, depending on the number of operations used for augmentation, we add additional units to accommodate the prediction of augmented classes. After training the models with LA, during the testing phase, we ignore the outputs for augmented labels—prior to the softmax operation—and only consider the class identity labels as the final output of the models. 
This is because the class categories in CIFAR and CIFAR-C datsets are not linked to augmentation operations like Plasma noise, etc. This could be thought of as asking a person to filter out what they see in a transformed picture and just identify the class without providing extra detail.

\textbf{Augmentation operations.} Corruptions employed in the CIFAR-C benchmark dataset~\citep{hendrycks2019benchmarking} include Gaussian Noise, Shot Noise, Impulse Noise, Defocus Blur, Glass Blur, Motion Blur, Zoom Blur, Snow, Frost, Fog, Brightness, Contrast, Elastic Transform, Pixelate, and JPEG Compression. According to~\citet{hendrycks2019benchmarking}, models evaluated on CIFAR-C should avoid using identical augmentations as those represented in the benchmark. 

AutoAugment~\citep{cubuk2019autoaugment} searches for various operations for data augmentation, as well as probabilities and magnitudes at which operations are applied. Through this, it identifies the optimal policy for models to achieve the highest validation accuracy on a given target dataset. The operations available for selection during the search, in AutoAugment, includes five geometric transforms (shear x/y, translate x/y, and rotate), two color transforms (color, invert), six intensity transforms (brightness, sharpness, solarize, equalize, autocontrast, contrast, and posterize), as well as cutout~\citep{devries2017improved} and sample pairing~\citep{zhang2017mixup, inoue2018data}. Some of these transformations may overlap with those in CIFAR-C.

To avoid this, AugMix~\citep{hendrycks2019augmix} integrates augmentations from AutoAugment, that do not overlap with the CIFAR-C benchmark~\citep{hendrycks2019benchmarking}. Specifically, AugMix employs five geometric transforms (shear x/y, translate x/y, and rotate) and four intensity transforms (solarize, equalize, autocontrast, and posterize). However, ensuring a complete independence between augmentations is challenging. For example,~\citep{rusak2020simple} and~\citep{mintun2021interaction} highlight the similarity between posterize and JPEG compression, as well as shear and translation to blur, respectively. 
Taking this into account, since we evaluate the robustness on CIFAR-C and also want to compare with AutoAugment and AugMix, we therefore, choose Plasma noise~\citep{nicolaou2022tormentor}, Planckian Jitter~\citep{zini2022planckian}, and Gamma adjustment as augmentations to be disjoint from the three. Noteworthy, for the sake of complete comparisons between methods, in our implementation of AutoAugment for CIFAR-C evaluation, unlike AugMix, we do not remove overlapping augmentation like cutout~\citep{devries2017improved}, brightness, etc. Moreover, we conduct additional experiments by incorporating Augmix as the augmentation operation in LA. We denote these experiments as AugMix$^{\mdoubleplus}$ in Tables~\ref{tab:allCF10} and~\ref{tab:allCF100}. In case of adversarial training, we employ FGSM~\citep{goodfellow2014explaining} and 10-step iterative PGD~\citep{madry2017towards} both with $L_\infty$ constraints and $\varepsilon= 0.3$. 

\textbf{Label Smoothing and Multi-Task Learning.} Label smoothing (LS) is an effective technique for regularizing DNNs. It achieves this by generating soft labels through a weighted average between a uniform distribution and the original hard labels. LS is typically used to address overfitting during training, leading to improved classification accuracy~\citep{szegedy2016rethinking, muller2019does}. While LS distributes the probability mass between the correct class label and all other classes, LA allocates the probability mass only between the identity class label and the augmentation labels. We compare the performance of LA against LS, where both techniques employ the same smoothing factor, $\delta$.
Furthermore, we compare LA with Multi-Task Learning (MTL), which offers an alternative approach to distinguishing between class identity and an indicator of augmentations. MTL involves optimizing a neural network across multiple related tasks simultaneously, aiming to enhance their generalization capabilities by leveraging shared patterns and representations among tasks~\citep{standley2020tasks,zhang2021survey, xin2022current}.
We use a 40-2 Wide ResNet architecture as a shared feature extractor, along with two task-specific output heads: one for predicting class identity and the other for identifying the type of distortions applied to images. We employ $\delta$ as the weight for the task of augmentation predictions and $1-\delta$ for the task of class identity prediction. The results of these experiments are presented in Table~\ref{tab:all_MTL_LS}, for both CIFAR datasets.

\subsection{Results}
\label{sec:results}
\textbf{Comparisons of regular data augmentation versus LA.
} Table~\ref{tab:cf10_LC} compares standard training—where the model is trained on clean data and tested on clean data—with training via LA when using different numbers of operations for augmentations.
The clean error improves when employing a single operation, while using two and three operations results in greater error reduction, up to 17.51\% when employing Plasma and Gamma ($i.e.$ $\text{P.G.$^{\mdoubleplus}$}$, where notation $^{\mdoubleplus}$ signifies label augmentation with factor $\delta$). Similarly, we observe improvement in mCE when employing three operations. Specifically, $\text{P.G.$^{\mdoubleplus}$}$ achieves a 22.70\% improvement compared to the mCE of standard training. Introducing one additional operation, Jitter, results in a further improvement of 28.56\% compared to the standard. Both calibration errors, ECE and RMS, show improvement when utilizing LA. However, values demonstrate no correlation between the number of operations employed and the improvement achieved. Similarly, errors under adversarial attacks FGSM and PGD improve, yet there is no apparent relationship between the number of operations employed and the observed enhancements. More details can be found in Fig.~\textcolor{red}{4}~\ref{fig:WRN_LA_vs_AUG}. 

The results from repeating the experiment without using LA are shown in Table~\ref{tab:cf10_No_LC}. In every instance, the Clean and calibration errors deteriorate. The mCE improves in most cases by up to 7.14\% when using three operations. The adversarial error shows both improvements and deteriorations in different  cases. Overall, there is no clear pattern between the number of operations and the fluctuation of errors.

Based on Tables~\ref{tab:cf10_LC} and~\ref{tab:cf10_No_LC}, as well as Fig.~\textcolor{red}{4}~\ref{fig:WRN_LA_vs_AUG}, 
it is evident that LA outperforms standard augmentation in minimizing Clean, mCE, calibration, and adversarial errors. As $\text{P.G.$^{\mdoubleplus}$}$ demonstrates the most significant enhancement in clean and PGD errors on Wide ResNet-50, while $\text{P.G.J.$^{\mdoubleplus}$}$ show better improvements in mCE and calibration, we proceed with these two operations to compare LA with other augmentation methods across different networks.

\textbf{Comparisons of other augmentations versus LA.}  Tables~\ref{tab:allCF10} and ~\ref{tab:allCF100} summarize the results obtained by applying Mixup~\citep{zhang2017mixup}, AugMix~\citep{hendrycks2019augmix}, AutoAugment~\citep{cubuk2019autoaugment}, adversarial training with FGSM~\citep{goodfellow2014explaining}, 10-step iterative PGD~\citep{madry2017towards}, and LA for various networks, for CIFAR-10 and CIFAR-100 datasets, respectively. 

\textbf{Robustness enhancement for CIFAR-10.} From the data presented in Table~\ref{tab:allCF10}, clearly LA contributes to an improvement in Clean error across various architectures. When considering the average performance across the five networks, LA outperforms 
all other methods,with $\text{P.G.J.$^{\mdoubleplus}$}$ improving the standard training by 23.29\%. In terms of enhancing robustness against common corruptions, there is a consistent improvement across AugMix, AutoAugment, FGSM and LA. However, depending on the employed networks, there is a variation in error rates for Mixup training, either showing a decrease or increase in comparison to the baseline. In all cases, PGD worsens both clean and mCE.

When comparing improvements in calibration errors, similar to the findings of~\citep{wang2023pitfall}, our results show that mixup training tends to increase calibration errors compared to standard training. Similarly, on average, adversarial training negatively affects ECE error. While RMS improves with FGSM training by 4.95\%, PGD diminishes it. Other methods all improve uncertainty estimates, among which, AutoAugment  outperforms others, with reductions of 44.33\% and 37.01\% in ECE and RMS, respectively. Among LA training experiments, AugMix$^{\mdoubleplus}$ surpasses other augmentations with reductions in ECE and RMS by 33.65\% and 31.15\%, while P.G.J.$^{\mdoubleplus}$ reduces the calibration errors for both metrics by 2.18\% and 7.65\%. 
Under both $\varepsilon$ budgets of 0.03 and 0.3, Mixup, FGSM, PGD, and all LA trainings show improvements. The enhancement in robustness from training with FGSM ($\varepsilon$=0.3) against FGSM attacks are 16.15\% and 47.22\%, compared to the standard adversarial error. However, training with FGSM barely improves error rates on PGD attack with $\varepsilon=0.3$. 
In contrast, LA not only generalizes to both attacks but also outperforms adversarial training in both FGSM and PGD when considering $\text{P.G.J.$^{\mdoubleplus}$}$. More specifically, the robustness to FGSM and PGD improves by 61.24\% and 49.54\% at budget $\varepsilon=0.03$ and 53.18\%  and 24.46\% at budget $\varepsilon=0.3$, respectively.
More details on the percentages of improvement compared to standard training can be found in Table~\ref{tab:percentage_cf10}.

\textbf{Robustness enhancement for CIFAR-100.}
Table~\ref{tab:allCF100} presents the performance of the aforementioned methods on the CIFAR-100 dataset.
Except for Mixup and adversarial training methods, the remaining techniques all improve clean errors, compare to standard training. Specifically, AugMix, AutoAugment, $\text{P.G.$^{\mdoubleplus}$}$, $\text{P.G.J.$^{\mdoubleplus}$}$, and $\text{AugMix$^{\mdoubleplus}$}$ reduce the error by 1.58\%, 4.49\%, 3.24\%, 4.31\%, and 3.53\%, respectively. In terms of calibration error, on average, almost all methods show improvements compared to standard training, except for PGD. For both low and high attack budgets, all methods improve the baseline except for AutoAugment and AugMix under PGD attacks. With $\varepsilon=0.03$, the robustness gain from LA training exceeds that from FGSM and PGD training. Specifically, FGSM and PGD reduce FGSM error by 9.87\% and 8.46\%, respectively, whereas $\text{P.G.J.$^{\mdoubleplus}$}$ reduces it by 26.94\%. Similarly, with $\varepsilon=0.03$, LA outperforms adversarial training. FGSM and PGD reduce PGD error by 7.94\% and 13.66\%, respectively, whereas $\text{P.G.J.$^{\mdoubleplus}$}$ reduces it by 18.41\%. However, with a higher budget of $\varepsilon=0.3$, the gains from FGSM and PGD training for FGSM attacks are 28.29\% and 24.40\%, while $\text{P.G.J.$^{\mdoubleplus}$}$ achieves a gain of 16.93\%. The robustness gains for PGD are 0.62\%, 10.03\%, and 10.38\% for FGSM, PGD, and $\text{P.G.J.$^{\mdoubleplus}$}$, respectively. More details can be found in Table~\ref{tab:percentage_cf100}.


\begin{table}[hbt!] 
\centering
\small
\renewcommand{\arraystretch}{0.82}
\setlength{\tabcolsep}{7pt} 
\begin{tabular}{llcccccccc}
\toprule
\multicolumn{2}{c}{\hspace{0.5cm}} & \multicolumn{8}{c}{Operations used } \\
\cmidrule{3-10}
\multicolumn{1}{l}{\hspace{0.5cm}} & \text{Error} & \text{Std.} & \text{P.$^{\mdoubleplus}$} & \text{G.$^{\mdoubleplus}$} & \text{J.$^{\mdoubleplus}$} & \text{P.G.$^{\mdoubleplus}$} & $\text{P.J.$^{\mdoubleplus}$}$ & $\text{G.J.$^{\mdoubleplus}$}$ & $\text{P.G.J.$^{\mdoubleplus}$}$ \\
\arrayrulecolor{black}\cmidrule(lr){1-10}
\textbf{} & Clean & 9.54 & 9.80 & 9.11 & 9.05 & 7.87 & 8.12 & 8.37 & 8.18\\
\multicolumn{1}{l}{} & mCE & 22.69 & 18.90 & 19.64 & 20.59 & 17.54 & 18.49 & 18.98 & 16.21\\
\arrayrulecolor{black!20}\cmidrule(lr){2-10}
 & ECE &6.33 & 6.22 & 5.87 &5.84 & 5.31 & 6.19 & 5.76 & 5.23 \\
 & RMS & 10.52 & 9.78 & 9.34 & 9.05 & 8.66 & 9.46 & 9.53 &8.72\\
\cmidrule(lr){2-10}
 & FGSM & 69.13 & 49.76 & 49.65 & 38.73 & 39.33 & 32.37 & 44.39 & 34.61\\
 & PGD$_{40}$ & 94.82 & 82.77 & 81.18 & 83.44 & 69.71 & 73.65 & 77.08 & 67.89\\
\arrayrulecolor{black}\bottomrule
\end{tabular}
\vspace{1pt}
\caption{Performance comparisons of baseline to Label Augmentation with different operations on CIFAR-10 using the Wide ResNet-50 architecture. {P.$^{\mdoubleplus}$}, {G.$^{\mdoubleplus}$}, and {J.$^{\mdoubleplus}$} refer to Plasma noise, Gamma Adjustment, and Planckian Jitter, respectively. The symbol $\mdoubleplus$ denotes the concatenation of labels with a factor of $\delta$ during these operations. Both adversarial training, FGSM and PGD use $L_\infty$ constraints with $\varepsilon= 0.3$.}

\label{tab:cf10_LC}
\end{table}
\begin{table}[hbt!] 
\centering
\small
\renewcommand{\arraystretch}{0.82}
\setlength{\tabcolsep}{7pt} 
\begin{tabular}{llcccccccc}
\toprule
\multicolumn{2}{c}{\hspace{0.5cm}} & \multicolumn{8}{c}{Operations used} \\
\cmidrule{3-10}
\multicolumn{1}{l}{\hspace{0.5cm}} & \text{Error} & \text{Std.} & \text{P.} & \text{G.} & \text{J.} & \text{P.G.} & \text{P.J.} & \text{G.J.} & \text{P.G.J.} \\
\cmidrule(lr){1-10}
\textbf{} & Clean &
 9.54 & 10.36 & 10.49 & 10.11 & 10.75 & 10.23 & 9.76 & 10.17 \\
\multicolumn{1}{l}{} 
& mCE & 22.69 & 21.66 & 22.21 & 22.98 & 21.25 & 21.71 & 23.11 & 21.07 \\
\arrayrulecolor{black!20}\cmidrule(lr){2-10}
\multirow{2}{*}{\rotatebox{90}{ }}
& ECE & 6.33 & 8.12 & 8.44 & 8.03 & 8.22 & 8.27 & 8.33 & 8.61 \\
 & RMS & 10.52 & 13.10 & 13.54 & 12.67 & 12.55 & 12.96 & 12.87 & 13.74 \\
\cmidrule(lr){2-10}
 & FGSM & 69.13 & 67.89 & 71.49 & 68.88 & 67.12 & 69.53 & 67.99 & 68.68 \\
 & PGD$_{40}$ & 94.82 & 93.25 & 93.73 & 93.59 & 93.91 & 93.88 & 94.12 & 93.90 \\
\arrayrulecolor{black}\bottomrule
\end{tabular}
\vspace{1pt}
\caption{The performance comparisons between the baseline and normal augmentation  with different operations on CIFAR-10 using the Wide ResNet-50 architecture. P., G., and J. represent Plasma noise, Gamma Adjustment, and Planckian Jitter, respectively.
Both adversarial training, FGSM and PGD use $L_\infty$ constraints with $\varepsilon= 0.3$.
}
\label{tab:cf10_No_LC}
\end{table}
%
\begin{table}[hbt!] 
\centering
\renewcommand{\arraystretch}{0.9} 
\setlength{\tabcolsep}{3pt} 
\begin{tabular}{lllccccccccc}
\toprule
\multicolumn{3}{c}{\multirow{2}{*}{}} & \multicolumn{6}{c}{\footnotesize Train} \\
\cmidrule{4-12}
\multicolumn{3}{c}{} & \footnotesize \text{Std.} & \footnotesize \text{Mixup} & \footnotesize \text{AugMix} & \footnotesize \text{AutoAug.} & \footnotesize \text{FGSM} &  \footnotesize \text{PGD$_{10}$} & \footnotesize \text{P.G.$^{\mdoubleplus}$} & \footnotesize \text{P.G.J.$^{\mdoubleplus}$} & \footnotesize \text{AugMix$^{\mdoubleplus}$}\\
\midrule
  &
\multirow{4}{*}{\footnotesize {Clean}} 
& \footnotesize LeNet & \footnotesize14.97\tiny\textcolor{gray!95}{$\pm${0.53}} & \footnotesize16.81\tiny\textcolor{gray!95}{$\pm${0.47}} & \footnotesize14.82\tiny\textcolor{gray!95}{$\pm${0.46}} & \footnotesize14.49\tiny\textcolor{gray!95}{$\pm${0.29}} & \footnotesize14.83\tiny\textcolor{gray!95}{$\pm${0.58}} & \footnotesize26.57\tiny\textcolor{gray!95}{$\pm${0.95}} & \footnotesize14.11\tiny\textcolor{gray!95}{$\pm${0.65}} & \footnotesize12.89\tiny\textcolor{gray!95}{$\pm${0.27}} & \footnotesize17.65\tiny\textcolor{gray!95}{$\pm${0.75}}\\
&& \footnotesize ResNet & \footnotesize10.06\tiny\textcolor{gray!95}{$\pm${0.22}} & \footnotesize10.43\tiny\textcolor{gray!95}{$\pm${0.58}} & \footnotesize9.87\tiny\textcolor{gray!95}{$\pm${0.11}} & \footnotesize9.12\tiny\textcolor{gray!95}{$\pm${0.25}} & \footnotesize10.59\tiny\textcolor{gray!95}{$\pm${0.36}} & \footnotesize21.84\tiny\textcolor{gray!95}{$\pm${0.73}} & \footnotesize8.89\tiny\textcolor{gray!95}{$\pm${0.14}} & \footnotesize8.46\tiny\textcolor{gray!95}{$\pm${0.25}} & \footnotesize9.53\tiny\textcolor{gray!95}{$\pm${0.25}}\\
\bottomrule
\end{tabular}
\caption{Error rates of various methods across different architectures for CIFAR-10 dataset. LA improves Clean and mCE compared to standard training, and also exhibits superior robustness to adversarial examples and and generalization to attacks, even outperforming adversarial training.}
\label{tab:allCF10}
\end{table}

\clearpage

\begin{table}[hbt!]
\centering
\renewcommand{\arraystretch}{0.9} 
\setlength{\tabcolsep}{3pt} 
\begin{tabular}{lllccccccccc}
\toprule
\multicolumn{3}{c}{\multirow{2}{*}{}} & \multicolumn{6}{c}{\footnotesize Train} \\
\cmidrule{4-12}
\multicolumn{3}{c}{} & \footnotesize \text{Std.} & \footnotesize \text{Mixup} & \footnotesize \text{AugMix} & \footnotesize \text{AutoAug.} & \footnotesize \text{FGSM} &  \footnotesize \text{PGD$_{10}$} & \footnotesize \text{P.G.$^{\mdoubleplus}$} & \footnotesize \text{P.G.J.$^{\mdoubleplus}$} & \footnotesize \text{AugMix$^{\mdoubleplus}$}\\
\midrule
  &
\multirow{4}{*}{\footnotesize {Clean}}
& \footnotesize LeNet & \footnotesize39.84\tiny\textcolor{gray!95}{$\pm${0.31}} & \footnotesize41.55\tiny\textcolor{gray!95}{$\pm${0.54}} & \footnotesize39.41\tiny\textcolor{gray!95}{$\pm${0.66}} & \footnotesize38.89\tiny\textcolor{gray!95}{$\pm${0.34}} & \footnotesize41.93\tiny\textcolor{gray!95}{$\pm${0.58}} & \footnotesize56.99\tiny\textcolor{gray!95}{$\pm${0.89}} & \footnotesize38.80\tiny\textcolor{gray!95}{$\pm${0.68}} & \footnotesize39.22\tiny\textcolor{gray!95}{$\pm${0.42}} & \footnotesize40.07\tiny\textcolor{gray!95}{$\pm${0.87}}\\
&& \footnotesize ResNet & \footnotesize31.10\tiny\textcolor{gray!95}{$\pm${0.25}} & \footnotesize32.24\tiny\textcolor{gray!95}{$\pm${0.53}} & \footnotesize30.87\tiny\textcolor{gray!95}{$\pm${0.52}} & \footnotesize29.91\tiny\textcolor{gray!95}{$\pm${0.32}} & \footnotesize33.97\tiny\textcolor{gray!95}{$\pm${0.36}} & \footnotesize47.47\tiny\textcolor{gray!95}{$\pm${0.63}} & \footnotesize30.67\tiny\textcolor{gray!95}{$\pm${0.61}} & \footnotesize30.15\tiny\textcolor{gray!95}{$\pm${0.47}} & \footnotesize29.72\tiny\textcolor{gray!95}{$\pm${0.62}}\\
\end{tabular}
\caption{Error rates of various methods across different architectures for the CIFAR-100 dataset. LA improves Clean, mCE, and adversarial errors  compared to standard training. Also, under lower-budget attacks, it consistently outperforms adversarial training.}
\label{tab:allCF100}
\end{table}

\clearpage

\textbf{Comparisons of LS and MTL versus LA.}
Table~\ref{tab:all_MTL_LS} presents a summary of the comparison between LS and MTL with LA. For CIFAR-10, the improvement in clean error is 5.22\%, 3.66\%, and 15.57\% for LS, MTL, and LA, respectively. The ECE increases by 32.96\% and 14.26\% for LS and MTL, respectively, whereas LA enhances both ECE and RMS calibration error by up to 1.90\%.

In both low and high $\varepsilon$ budget attacks, LA exhibits superior improvement compared to LS and MTL, with enhancements of up to 50.29\% and 28.27\% for FGSM and PGD with $\varepsilon=0.3$, respectively. Nevertheless, the effectiveness of LS in improving adversarial error for CIFAR-10 data is noticeable, and have been highlighted in  findings of~\citep{shafahi2019label, pang2020bag,ren2021adversarial}, as well.

Similarly, in CIFAR-100, LA outperforms in mCE improvement with 8.32\% error reduction. In terms of adversarial errors, it improves the baseline by 29.87\% for FGSM and 25.11\% for PGD under a low budget $\varepsilon=0.03$, as well as by 20.59\% and 11.93\% with a higher budget of $\varepsilon=0.3$ under FGSM and PGD attacks, respectively.

\begin{table}[h]
\centering
\small 
\renewcommand{\arraystretch}{1.3} 
\setlength{\tabcolsep}{3pt} 
\begin{minipage}[t]{0.45\linewidth}
\centering
\begin{tabular}{@{}clcccc@{}}
\toprule
\small & & \multicolumn{3}{c}{\small Train} \\ \cmidrule(lr){3-6}
\small & Err. & \small Std.  & \small LS. & \small MTL. & \small \text{P.G.J.$^{\mdoubleplus}$}\\
\midrule
& Clean 
&\footnotesize9.57\tiny\textcolor{gray!95}{$\pm${0.21}}  
&\footnotesize 9.07\tiny\textcolor{gray!95}{$\pm${1.25}}
&\footnotesize 9.22\tiny\textcolor{gray!95}{$\pm${0.23}}
&\footnotesize 8.08\tiny\textcolor{gray!95}{$\pm${0.19}}
\\
& mCE 
&\footnotesize22.73\tiny\textcolor{gray!95}{$\pm${0.27}} &\footnotesize21.89\tiny\textcolor{gray!95}{$\pm${1.79}} &\footnotesize20.59\tiny\textcolor{gray!95}{$\pm${0.58}} &\footnotesize16.78\tiny\textcolor{gray!95}{$\pm${0.71}}\\
\arrayrulecolor{black!20}\cmidrule{2-6}
& ECE  
&\footnotesize6.31\tiny\textcolor{gray!95}{$\pm${0.21}} &\footnotesize8.39\tiny\textcolor{gray!95}{$\pm${2.03}} &\footnotesize7.21\tiny\textcolor{gray!95}{$\pm${0.81}} &\footnotesize6.19\tiny\textcolor{gray!95}{$\pm${0.26}}\\
& RMS 
&\footnotesize10.37\tiny\textcolor{gray!95}{$\pm${0.38}} &\footnotesize10.92\tiny\textcolor{gray!95}{$\pm${1.88}} &\footnotesize11.17\tiny\textcolor{gray!95}{$\pm${0.25}} &\footnotesize8.96\tiny\textcolor{gray!95}{$\pm${0.43}}\\
\arrayrulecolor{black!20}\cmidrule{2-6}
\multirow{2}{*}{\rotatebox{90}{\small \parbox{1.1cm}{\centering \tiny$\varepsilon \tiny= \tiny0.03$}}}
& FGSM 
&\footnotesize43.14\tiny\textcolor{gray!95}{$\pm${0.23}} &\footnotesize27.01\tiny\textcolor{gray!95}{$\pm${1.02}} &\footnotesize41.31\tiny\textcolor{gray!95}{$\pm${0.19}} &\footnotesize14.44\tiny\textcolor{gray!95}{$\pm${0.70}}\\
& PGD 
&\footnotesize77.08\tiny\textcolor{gray!95}{$\pm${0.18}} &\footnotesize56.6\tiny\textcolor{gray!95}{$\pm${1.21}} &\footnotesize66.06\tiny\textcolor{gray!95}{$\pm${0.62}} &\footnotesize37.81\tiny\textcolor{gray!95}{$\pm${0.37}}\\
\arrayrulecolor{black!20}\cmidrule{2-6}
\multirow{2}{*}{\rotatebox{90}{\small \parbox{1.1cm}{\centering \tiny$\varepsilon \tiny= \tiny0.3$}}}
& FGSM 
&\footnotesize69.53\tiny\textcolor{gray!95}{$\pm${0.88}} &\footnotesize52.36\tiny\textcolor{gray!95}{$\pm${2.66}} &\footnotesize71.38\tiny\textcolor{gray!95}{$\pm${0.55}} &\footnotesize34.56\tiny\textcolor{gray!95}{$\pm${0.38}}\\
& PGD 
&\footnotesize94.78\tiny\textcolor{gray!95}{$\pm${0.53}} &\footnotesize82.43\tiny\textcolor{gray!95}{$\pm${1.54}} &\footnotesize93.36\tiny\textcolor{gray!95}{$\pm${0.73}} &\footnotesize67.99\tiny\textcolor{gray!95}{$\pm${1.53}}\\
%
%
\arrayrulecolor{black}\cmidrule{2-6}
\end{tabular}
\textbf{CIFAR-10}
\end{minipage}
\begin{minipage}[t]{0.5\linewidth}
\centering
\setlength{\tabcolsep}{3pt} 
\begin{tabular}{@{}clcccc@{}}
\toprule
\small & & \multicolumn{3}{c}{\small Train} \\ 
\cmidrule(lr){3-6}
\small & Err. & \small Std.  & \small LS. & \small MTL. & \small \text{P.G.J.$^{\mdoubleplus}$}\\
\midrule
& Clean
&\footnotesize29.43\tiny\textcolor{gray!95}{$\pm${0.19}} &\footnotesize27.85\tiny\textcolor{gray!95}{$\pm${1.38}} &\footnotesize32.23\tiny\textcolor{gray!95}{$\pm${0.45}} &\footnotesize28.59\tiny\textcolor{gray!95}{$\pm${0.31}}\\
& mCE 
&\footnotesize48.33\tiny\textcolor{gray!95}{$\pm${0.25}} &\footnotesize47.11\tiny\textcolor{gray!95}{$\pm${1.91}} &\footnotesize48.32\tiny\textcolor{gray!95}{$\pm${0.56}} &\footnotesize44.31\tiny\textcolor{gray!95}{$\pm${0.39}}\\
\arrayrulecolor{black!20}\cmidrule{2-6}
& ECE 
&\footnotesize18.51\tiny\textcolor{gray!95}{$\pm${0.33}} &\footnotesize7.89\tiny\textcolor{gray!95}{$\pm${1.83}} &\footnotesize22.26\tiny\textcolor{gray!95}{$\pm${0.92}} &\footnotesize16.41\tiny\textcolor{gray!95}{$\pm${0.39}}\\
& RMS
&\footnotesize22.94\tiny\textcolor{gray!95}{$\pm${0.46}} &\footnotesize8.78\tiny\textcolor{gray!95}{$\pm${1.45}} &\footnotesize26.79\tiny\textcolor{gray!95}{$\pm${0.33}} &\footnotesize18.15\tiny\textcolor{gray!95}{$\pm${0.43}}\\
\arrayrulecolor{black!20}\cmidrule{2-6}
\multirow{2}{*}{\rotatebox{90}{\small \parbox{1.1cm}{\centering \tiny$\varepsilon \tiny= \tiny0.03$}}}
& FGSM 
&\footnotesize61.97\tiny\textcolor{gray!95}{$\pm${0.42}} &\footnotesize53.21\tiny\textcolor{gray!95}{$\pm${0.98}} &\footnotesize62.01\tiny\textcolor{gray!95}{$\pm${0.36}} &\footnotesize43.46\tiny\textcolor{gray!95}{$\pm${0.62}}\\
& PGD
&\footnotesize82.01\tiny\textcolor{gray!95}{$\pm${0.35}} &\footnotesize76.22\tiny\textcolor{gray!95}{$\pm${1.10}} &\footnotesize77.06\tiny\textcolor{gray!95}{$\pm${0.74}} &\footnotesize61.42\tiny\textcolor{gray!95}{$\pm${0.77}}\\
\arrayrulecolor{black!20}\cmidrule{2-6}
\multirow{2}{*}{\rotatebox{90}{\small \parbox{1.1cm}{\centering \tiny$\varepsilon \tiny= \tiny0.3$}}}
& FGSM
&\footnotesize85.46\tiny\textcolor{gray!95}{$\pm${0.82}} &\footnotesize80.44\tiny\textcolor{gray!95}{$\pm${1.55}} &\footnotesize84.04\tiny\textcolor{gray!95}{$\pm${0.92}} &\footnotesize67.86\tiny\textcolor{gray!95}{$\pm${0.79}}\\
& PGD
&\footnotesize93.11\tiny\textcolor{gray!95}{$\pm${0.79}} &\footnotesize91.66\tiny\textcolor{gray!95}{$\pm${2.03}} &\footnotesize90.16\tiny\textcolor{gray!95}{$\pm${0.65}} &\footnotesize82.00\tiny\textcolor{gray!95}{$\pm${0.68}}\\
%
%
%
\arrayrulecolor{black}\cmidrule{2-6}
\end{tabular}
\textbf{CIFAR-100}
\end{minipage}
\caption{Performance comparisons of Label Smoothing and Multi-task Learning to  LA ({P.G.J.$^{\mdoubleplus}$}) using the Wide ResNet-50 architecture. For both CIFAR-10 and CIFAR-100, LA improves Clean, mCE, Calibration, and adversarial errors compared to standard training and consistently outperforms LS and MTL in improving adversarial robustness.}
\label{tab:all_MTL_LS}
\end{table}

\section{Conclusion}

To align our naming convention and label assignment when training the DNNs, we developed Label Augmentation. Essentially, LA assigns one-hot labels to each of the operations used during augmentations. Then, instead of solely augmenting transformed data in the training pipeline, LA involves augmenting labels by concatenating input labels with operation labels, using a factor of $\delta$. This automatically enriches the labels without requiring extra human annotation and has proved to be advantageous in enhancing both robustness against common and adversarial perturbations. In terms of Clean and mCE error, comparative analysis shows LA performing nearly as well as AugMix and AutoAugment. However, in terms of adversarial robustness, LA is significantly better than other augmentation methods and can even outperform adversarial training. 
LA is flexible and could be employed in other modalities. For instance, future works can utilize LA in audio inputs while training with noisy audio signals. 
This study, alongside much of the existing research on distributional shift, primarily focuses on evaluating model robustness to 2D image transformations, largely overlooking changes in viewpoint within 3D transformations found in various real-world applications (e.g., autonomous driving). It has been demonstrated that common image classifiers are highly vulnerable to adversarial viewpoints~\citep{dong2022viewfool}. 
Future studies could explore whether employing LA—with rotation as augmentation—can enhance robustness against adversarial viewpoints.

\section{Acknowledgement}

This work was conducted with the financial support of the Science
Foundation Ireland Centre for Research Training in Artificial Intelligence
under Grant No. 18/CRT/6223. We would also like to thank the anonymous reviewers for their helpful and informative comments.

\bibliography{collas2024_conference}

\begin{thebibliography}{85}
\providecommand{\natexlab}[1]{#1}
\providecommand{\url}[1]{\texttt{#1}}
\expandafter\ifx\csname urlstyle\endcsname\relax
  \providecommand{\doi}[1]{doi: #1}\else
  \providecommand{\doi}{doi: \begingroup \urlstyle{rm}\Url}\fi

\bibitem[Andriushchenko et~al.(2020)Andriushchenko, Croce, Flammarion, and Hein]{andriushchenko2020square}
Maksym Andriushchenko, Francesco Croce, Nicolas Flammarion, and Matthias Hein.
\newblock Square attack: a query-efficient black-box adversarial attack via random search.
\newblock In \emph{European conference on computer vision}, pp.\  484--501. Springer, 2020.

\bibitem[Athalye et~al.(2018)Athalye, Carlini, and Wagner]{athalye2018obfuscated}
Anish Athalye, Nicholas Carlini, and David Wagner.
\newblock Obfuscated gradients give a false sense of security: Circumventing defenses to adversarial examples.
\newblock In \emph{International conference on machine learning}, pp.\  274--283. PMLR, 2018.

\bibitem[Azulay \& Weiss(2018)Azulay and Weiss]{azulay2018deep}
Aharon Azulay and Yair Weiss.
\newblock Why do deep convolutional networks generalize so poorly to small image transformations?
\newblock \emph{arXiv preprint arXiv:1805.12177}, 2018.

\bibitem[Bai et~al.(2021)Bai, Luo, Zhao, Wen, and Wang]{bai2021recent}
Tao Bai, Jinqi Luo, Jun Zhao, Bihan Wen, and Qian Wang.
\newblock Recent advances in adversarial training for adversarial robustness.
\newblock \emph{arXiv preprint arXiv:2102.01356}, 2021.

\bibitem[Bengio et~al.(2013)Bengio, Courville, and Vincent]{bengio2013representation}
Yoshua Bengio, Aaron Courville, and Pascal Vincent.
\newblock Representation learning: A review and new perspectives.
\newblock \emph{IEEE transactions on pattern analysis and machine intelligence}, 35\penalty0 (8):\penalty0 1798--1828, 2013.

\bibitem[Bu et~al.(2023)Bu, Huang, and Cui]{bu2023towards}
Qingwen Bu, Dong Huang, and Heming Cui.
\newblock Towards building more robust models with frequency bias.
\newblock In \emph{Proceedings of the IEEE/CVF International Conference on Computer Vision}, pp.\  4402--4411, 2023.

\bibitem[Cai et~al.(2023)Cai, Ning, Yang, and Wang]{cai2023ensemble}
Yi~Cai, Xuefei Ning, Huazhong Yang, and Yu~Wang.
\newblock Ensemble-in-one: ensemble learning within random gated networks for enhanced adversarial robustness.
\newblock In \emph{Proceedings of the AAAI Conference on Artificial Intelligence}, volume~37, pp.\  14738--14747, 2023.

\bibitem[Carmon et~al.(2019)Carmon, Raghunathan, Schmidt, Duchi, and Liang]{carmon2019unlabeled}
Yair Carmon, Aditi Raghunathan, Ludwig Schmidt, John~C Duchi, and Percy~S Liang.
\newblock Unlabeled data improves adversarial robustness.
\newblock \emph{Advances in neural information processing systems}, 32, 2019.

\bibitem[Croce \& Hein(2020{\natexlab{a}})Croce and Hein]{croce2020minimally}
Francesco Croce and Matthias Hein.
\newblock Minimally distorted adversarial examples with a fast adaptive boundary attack.
\newblock In \emph{International Conference on Machine Learning}, pp.\  2196--2205. PMLR, 2020{\natexlab{a}}.

\bibitem[Croce \& Hein(2020{\natexlab{b}})Croce and Hein]{croce2020reliable}
Francesco Croce and Matthias Hein.
\newblock Reliable evaluation of adversarial robustness with an ensemble of diverse parameter-free attacks.
\newblock In \emph{International conference on machine learning}, pp.\  2206--2216. PMLR, 2020{\natexlab{b}}.

\bibitem[Cubuk et~al.(2019)Cubuk, Zoph, Mane, Vasudevan, and Le]{cubuk2019autoaugment}
Ekin~D Cubuk, Barret Zoph, Dandelion Mane, Vijay Vasudevan, and Quoc~V Le.
\newblock Autoaugment: Learning augmentation strategies from data.
\newblock In \emph{Proceedings of the IEEE/CVF conference on computer vision and pattern recognition}, pp.\  113--123, 2019.

\bibitem[Deng et~al.(2021{\natexlab{a}})Deng, Yang, Xu, Su, and Zhu]{deng2021libre}
Zhijie Deng, Xiao Yang, Shizhen Xu, Hang Su, and Jun Zhu.
\newblock Libre: A practical bayesian approach to adversarial detection.
\newblock In \emph{Proceedings of the IEEE/CVF conference on computer vision and pattern recognition}, pp.\  972--982, 2021{\natexlab{a}}.

\bibitem[Deng et~al.(2021{\natexlab{b}})Deng, Zhang, Ghorbani, and Zou]{deng2021improving}
Zhun Deng, Linjun Zhang, Amirata Ghorbani, and James Zou.
\newblock Improving adversarial robustness via unlabeled out-of-domain data.
\newblock In \emph{International Conference on Artificial Intelligence and Statistics}, pp.\  2845--2853. PMLR, 2021{\natexlab{b}}.

\bibitem[DeVries \& Taylor(2017)DeVries and Taylor]{devries2017improved}
Terrance DeVries and Graham~W Taylor.
\newblock Improved regularization of convolutional neural networks with cutout.
\newblock \emph{arXiv preprint arXiv:1708.04552}, 2017.

\bibitem[Dodge \& Karam(2016)Dodge and Karam]{dodge2016understanding}
Samuel Dodge and Lina Karam.
\newblock Understanding how image quality affects deep neural networks.
\newblock In \emph{2016 eighth international conference on quality of multimedia experience (QoMEX)}, pp.\  1--6. IEEE, 2016.

\bibitem[Dong et~al.(2022)Dong, Ruan, Su, Kang, Wei, and Zhu]{dong2022viewfool}
Yinpeng Dong, Shouwei Ruan, Hang Su, Caixin Kang, Xingxing Wei, and Jun Zhu.
\newblock Viewfool: Evaluating the robustness of visual recognition to adversarial viewpoints.
\newblock \emph{Advances in Neural Information Processing Systems}, 35:\penalty0 36789--36803, 2022.

\bibitem[Eom \& Ham(2019)Eom and Ham]{eom2019learning}
Chanho Eom and Bumsub Ham.
\newblock Learning disentangled representation for robust person re-identification.
\newblock \emph{Advances in neural information processing systems}, 32, 2019.

\bibitem[Ford et~al.(2019)Ford, Gilmer, Carlini, and Cubuk]{ford2019adversarial}
Nic Ford, Justin Gilmer, Nicolas Carlini, and Dogus Cubuk.
\newblock Adversarial examples are a natural consequence of test error in noise.
\newblock \emph{arXiv preprint arXiv:1901.10513}, 2019.

\bibitem[Gabbay \& Hoshen(2019)Gabbay and Hoshen]{gabbay2019demystifying}
Aviv Gabbay and Yedid Hoshen.
\newblock Demystifying inter-class disentanglement.
\newblock \emph{arXiv preprint arXiv:1906.11796}, 2019.

\bibitem[Gawlikowski et~al.(2021)Gawlikowski, Tassi, Ali, Lee, Humt, Feng, Kruspe, Triebel, Jung, Roscher, et~al.]{gawlikowski2021survey}
Jakob Gawlikowski, Cedrique Rovile~Njieutcheu Tassi, Mohsin Ali, Jongseok Lee, Matthias Humt, Jianxiang Feng, Anna Kruspe, Rudolph Triebel, Peter Jung, Ribana Roscher, et~al.
\newblock A survey of uncertainty in deep neural networks.
\newblock \emph{arXiv preprint arXiv:2107.03342}, 2021.

\bibitem[Geirhos et~al.(2018)Geirhos, Rubisch, Michaelis, Bethge, Wichmann, and Brendel]{geirhos2018imagenet}
Robert Geirhos, Patricia Rubisch, Claudio Michaelis, Matthias Bethge, Felix~A Wichmann, and Wieland Brendel.
\newblock Imagenet-trained cnns are biased towards texture; increasing shape bias improves accuracy and robustness.
\newblock \emph{arXiv preprint arXiv:1811.12231}, 2018.

\bibitem[Goodfellow et~al.(2014)Goodfellow, Shlens, and Szegedy]{goodfellow2014explaining}
Ian~J Goodfellow, Jonathon Shlens, and Christian Szegedy.
\newblock Explaining and harnessing adversarial examples.
\newblock \emph{arXiv preprint arXiv:1412.6572}, 2014.

\bibitem[Gowal et~al.(2021)Gowal, Rebuffi, Wiles, Stimberg, Calian, and Mann]{gowal2021improving}
Sven Gowal, Sylvestre-Alvise Rebuffi, Olivia Wiles, Florian Stimberg, Dan~Andrei Calian, and Timothy~A Mann.
\newblock Improving robustness using generated data.
\newblock \emph{Advances in Neural Information Processing Systems}, 34:\penalty0 4218--4233, 2021.

\bibitem[Guo et~al.(2017)Guo, Pleiss, Sun, and Weinberger]{guo2017calibration}
Chuan Guo, Geoff Pleiss, Yu~Sun, and Kilian~Q Weinberger.
\newblock On calibration of modern neural networks.
\newblock In \emph{International conference on machine learning}, pp.\  1321--1330. PMLR, 2017.

\bibitem[HarryKim(2020)]{kim2020torchattacks}
HarryKim.
\newblock Torchattacks a pytorch library for adversarial attacks, 2020.
\newblock URL \url{https://adversarial-attacks-pytorch.readthedocs.io/en/latest/}.
\newblock Accessed: 2024-05-23.

\bibitem[He et~al.(2016)He, Zhang, Ren, and Sun]{he2016deep}
Kaiming He, Xiangyu Zhang, Shaoqing Ren, and Jian Sun.
\newblock Deep residual learning for image recognition.
\newblock In \emph{Proceedings of the IEEE conference on computer vision and pattern recognition}, pp.\  770--778, 2016.

\bibitem[Hendrycks \& Dietterich(2019)Hendrycks and Dietterich]{hendrycks2019benchmarking}
Dan Hendrycks and Thomas Dietterich.
\newblock Benchmarking neural network robustness to common corruptions and perturbations.
\newblock \emph{arXiv preprint arXiv:1903.12261}, 2019.

\bibitem[Hendrycks et~al.(2019)Hendrycks, Mu, Cubuk, Zoph, Gilmer, and Lakshminarayanan]{hendrycks2019augmix}
Dan Hendrycks, Norman Mu, Ekin~D Cubuk, Barret Zoph, Justin Gilmer, and Balaji Lakshminarayanan.
\newblock Augmix: A simple data processing method to improve robustness and uncertainty.
\newblock \emph{arXiv preprint arXiv:1912.02781}, 2019.

\bibitem[Hong et~al.(2021)Hong, Choi, and Kim]{hong2021stylemix}
Minui Hong, Jinwoo Choi, and Gunhee Kim.
\newblock Stylemix: Separating content and style for enhanced data augmentation.
\newblock In \emph{Proceedings of the IEEE/CVF conference on computer vision and pattern recognition}, pp.\  14862--14870, 2021.

\bibitem[Inoue(2018)]{inoue2018data}
Hiroshi Inoue.
\newblock Data augmentation by pairing samples for images classification.
\newblock \emph{arXiv preprint arXiv:1801.02929}, 2018.

\bibitem[Jackson et~al.(2019)Jackson, Abarghouei, Bonner, Breckon, and Obara]{jackson2019style}
Philip~TG Jackson, Amir~Atapour Abarghouei, Stephen Bonner, Toby~P Breckon, and Boguslaw Obara.
\newblock Style augmentation: data augmentation via style randomization.
\newblock In \emph{CVPR workshops}, volume~6, pp.\  10--11, 2019.

\bibitem[Kireev et~al.(2022)Kireev, Andriushchenko, and Flammarion]{kireev2022effectiveness}
Klim Kireev, Maksym Andriushchenko, and Nicolas Flammarion.
\newblock On the effectiveness of adversarial training against common corruptions.
\newblock In \emph{Uncertainty in Artificial Intelligence}, pp.\  1012--1021. PMLR, 2022.

\bibitem[Krizhevsky et~al.(2009)Krizhevsky, Hinton, et~al.]{krizhevsky2009learning}
Alex Krizhevsky, Geoffrey Hinton, et~al.
\newblock Learning multiple layers of features from tiny images.
\newblock 2009.

\bibitem[Krizhevsky et~al.(2012)Krizhevsky, Sutskever, and Hinton]{krizhevsky2012imagenet}
Alex Krizhevsky, Ilya Sutskever, and Geoffrey~E Hinton.
\newblock Imagenet classification with deep convolutional neural networks.
\newblock \emph{Advances in neural information processing systems}, 25, 2012.

\bibitem[Kurakin et~al.(2016)Kurakin, Goodfellow, and Bengio]{kurakin2016adversarial}
Alexey Kurakin, Ian Goodfellow, and Samy Bengio.
\newblock Adversarial machine learning at scale.
\newblock \emph{arXiv preprint arXiv:1611.01236}, 2016.

\bibitem[Lakshminarayanan et~al.(2017)Lakshminarayanan, Pritzel, and Blundell]{lakshminarayanan2017simple}
Balaji Lakshminarayanan, Alexander Pritzel, and Charles Blundell.
\newblock Simple and scalable predictive uncertainty estimation using deep ensembles.
\newblock \emph{Advances in neural information processing systems}, 30, 2017.

\bibitem[Laugros et~al.(2019)Laugros, Caplier, and Ospici]{laugros2019adversarial}
Alfred Laugros, Alice Caplier, and Matthieu Ospici.
\newblock Are adversarial robustness and common perturbation robustness independant attributes?
\newblock In \emph{Proceedings of the IEEE/CVF International Conference on Computer Vision Workshops}, pp.\  0--0, 2019.

\bibitem[LeCun et~al.(1998)LeCun, Bottou, Bengio, and Haffner]{lecun1998gradient}
Yann LeCun, L{\'e}on Bottou, Yoshua Bengio, and Patrick Haffner.
\newblock Gradient-based learning applied to document recognition.
\newblock \emph{Proceedings of the IEEE}, 86\penalty0 (11):\penalty0 2278--2324, 1998.

\bibitem[Li et~al.(2022)Li, Wu, Chen, Fang, and Huang]{li2022subspace}
Tao Li, Yingwen Wu, Sizhe Chen, Kun Fang, and Xiaolin Huang.
\newblock Subspace adversarial training.
\newblock In \emph{Proceedings of the IEEE/CVF Conference on Computer Vision and Pattern Recognition}, pp.\  13409--13418, 2022.

\bibitem[Lightning(2024)]{torchmetrics}
PyTorch Lightning.
\newblock {torchmetrics}: Metrics for pytorch.
\newblock \url{https://lightning.ai/docs/torchmetrics/stable/classification/calibration_error.html}, 2024.
\newblock Accessed: April 5, 2024.

\bibitem[Liu et~al.(2021{\natexlab{a}})Liu, Shen, He, Zhang, Xu, Yu, and Cui]{liu2021towards}
Jiashuo Liu, Zheyan Shen, Yue He, Xingxuan Zhang, Renzhe Xu, Han Yu, and Peng Cui.
\newblock Towards out-of-distribution generalization: A survey.
\newblock \emph{arXiv preprint arXiv:2108.13624}, 2021{\natexlab{a}}.

\bibitem[Liu et~al.(2021{\natexlab{b}})Liu, Lin, Cao, Hu, Wei, Zhang, Lin, and Guo]{liu2021swin}
Ze~Liu, Yutong Lin, Yue Cao, Han Hu, Yixuan Wei, Zheng Zhang, Stephen Lin, and Baining Guo.
\newblock Swin transformer: Hierarchical vision transformer using shifted windows.
\newblock In \emph{Proceedings of the IEEE/CVF international conference on computer vision}, pp.\  10012--10022, 2021{\natexlab{b}}.

\bibitem[Lopes et~al.(2019)Lopes, Yin, Poole, Gilmer, and Cubuk]{lopes2019improving}
Raphael~Gontijo Lopes, Dong Yin, Ben Poole, Justin Gilmer, and Ekin~D Cubuk.
\newblock Improving robustness without sacrificing accuracy with patch gaussian augmentation.
\newblock \emph{arXiv preprint arXiv:1906.02611}, 2019.

\bibitem[Loshchilov \& Hutter(2016)Loshchilov and Hutter]{loshchilov2016sgdr}
Ilya Loshchilov and Frank Hutter.
\newblock Sgdr: Stochastic gradient descent with warm restarts.
\newblock \emph{arXiv preprint arXiv:1608.03983}, 2016.

\bibitem[Lukasik et~al.(2020)Lukasik, Bhojanapalli, Menon, and Kumar]{lukasik2020does}
Michal Lukasik, Srinadh Bhojanapalli, Aditya Menon, and Sanjiv Kumar.
\newblock Does label smoothing mitigate label noise?
\newblock In \emph{International Conference on Machine Learning}, pp.\  6448--6458. PMLR, 2020.

\bibitem[Madry et~al.(2017)Madry, Makelov, Schmidt, Tsipras, and Vladu]{madry2017towards}
Aleksander Madry, Aleksandar Makelov, Ludwig Schmidt, Dimitris Tsipras, and Adrian Vladu.
\newblock Towards deep learning models resistant to adversarial attacks.
\newblock \emph{arXiv preprint arXiv:1706.06083}, 2017.

\bibitem[Metzen et~al.(2017)Metzen, Genewein, Fischer, and Bischoff]{metzen2017detecting}
Jan~Hendrik Metzen, Tim Genewein, Volker Fischer, and Bastian Bischoff.
\newblock On detecting adversarial perturbations.
\newblock \emph{arXiv preprint arXiv:1702.04267}, 2017.

\bibitem[Mintun et~al.(2021)Mintun, Kirillov, and Xie]{mintun2021interaction}
Eric Mintun, Alexander Kirillov, and Saining Xie.
\newblock On interaction between augmentations and corruptions in natural corruption robustness.
\newblock \emph{Advances in Neural Information Processing Systems}, 34:\penalty0 3571--3583, 2021.

\bibitem[Modas et~al.(2022)Modas, Rade, Ortiz-Jim{\'e}nez, Moosavi-Dezfooli, and Frossard]{modas2022prime}
Apostolos Modas, Rahul Rade, Guillermo Ortiz-Jim{\'e}nez, Seyed-Mohsen Moosavi-Dezfooli, and Pascal Frossard.
\newblock Prime: A few primitives can boost robustness to common corruptions.
\newblock In \emph{European Conference on Computer Vision}, pp.\  623--640. Springer, 2022.

\bibitem[Moosavi-Dezfooli et~al.(2016)Moosavi-Dezfooli, Fawzi, and Frossard]{moosavi2016deepfool}
Seyed-Mohsen Moosavi-Dezfooli, Alhussein Fawzi, and Pascal Frossard.
\newblock Deepfool: a simple and accurate method to fool deep neural networks.
\newblock In \emph{Proceedings of the IEEE conference on computer vision and pattern recognition}, pp.\  2574--2582, 2016.

\bibitem[M{\"u}ller et~al.(2019)M{\"u}ller, Kornblith, and Hinton]{muller2019does}
Rafael M{\"u}ller, Simon Kornblith, and Geoffrey~E Hinton.
\newblock When does label smoothing help?
\newblock \emph{Advances in neural information processing systems}, 32, 2019.

\bibitem[Nicolaou et~al.(2022)Nicolaou, Christlein, Riba, Shi, Vogeler, and Seuret]{nicolaou2022tormentor}
Anguelos Nicolaou, Vincent Christlein, Edgar Riba, Jian Shi, Georg Vogeler, and Mathias Seuret.
\newblock Tormentor: Deterministic dynamic-path, data augmentations with fractals.
\newblock In \emph{Proceedings of the IEEE/CVF Conference on Computer Vision and Pattern Recognition}, pp.\  2707--2711, 2022.

\bibitem[Ovadia et~al.(2019)Ovadia, Fertig, Ren, Nado, Sculley, Nowozin, Dillon, Lakshminarayanan, and Snoek]{ovadia2019can}
Yaniv Ovadia, Emily Fertig, Jie Ren, Zachary Nado, David Sculley, Sebastian Nowozin, Joshua Dillon, Balaji Lakshminarayanan, and Jasper Snoek.
\newblock Can you trust your model's uncertainty? evaluating predictive uncertainty under dataset shift.
\newblock \emph{Advances in neural information processing systems}, 32, 2019.

\bibitem[Pang et~al.(2018)Pang, Du, Dong, and Zhu]{pang2018towards}
Tianyu Pang, Chao Du, Yinpeng Dong, and Jun Zhu.
\newblock Towards robust detection of adversarial examples.
\newblock \emph{Advances in neural information processing systems}, 31, 2018.

\bibitem[Pang et~al.(2020)Pang, Yang, Dong, Su, and Zhu]{pang2020bag}
Tianyu Pang, Xiao Yang, Yinpeng Dong, Hang Su, and Jun Zhu.
\newblock Bag of tricks for adversarial training.
\newblock \emph{arXiv preprint arXiv:2010.00467}, 2020.

\bibitem[Papernot et~al.(2016{\natexlab{a}})Papernot, McDaniel, and Goodfellow]{papernot2016transferability}
Nicolas Papernot, Patrick McDaniel, and Ian Goodfellow.
\newblock Transferability in machine learning: from phenomena to black-box attacks using adversarial samples.
\newblock \emph{arXiv preprint arXiv:1605.07277}, 2016{\natexlab{a}}.

\bibitem[Papernot et~al.(2016{\natexlab{b}})Papernot, McDaniel, Wu, Jha, and Swami]{papernot2016distillation}
Nicolas Papernot, Patrick McDaniel, Xi~Wu, Somesh Jha, and Ananthram Swami.
\newblock Distillation as a defense to adversarial perturbations against deep neural networks.
\newblock In \emph{2016 IEEE symposium on security and privacy (SP)}, pp.\  582--597. IEEE, 2016{\natexlab{b}}.

\bibitem[Papernot et~al.(2018)Papernot, Faghri, Carlini, Goodfellow, Feinman, Kurakin, Xie, Sharma, Brown, Roy, Matyasko, Behzadan, Hambardzumyan, Zhang, Juang, Li, Sheatsley, Garg, Uesato, Gierke, Dong, Berthelot, Hendricks, Rauber, and Long]{papernot2018cleverhans}
Nicolas Papernot, Fartash Faghri, Nicholas Carlini, Ian Goodfellow, Reuben Feinman, Alexey Kurakin, Cihang Xie, Yash Sharma, Tom Brown, Aurko Roy, Alexander Matyasko, Vahid Behzadan, Karen Hambardzumyan, Zhishuai Zhang, Yi-Lin Juang, Zhi Li, Ryan Sheatsley, Abhibhav Garg, Jonathan Uesato, Willi Gierke, Yinpeng Dong, David Berthelot, Paul Hendricks, Jonas Rauber, and Rujun Long.
\newblock Technical report on the cleverhans v2.1.0 adversarial examples library.
\newblock \emph{arXiv preprint arXiv:1610.00768}, 2018.

\bibitem[Qin et~al.(2022)Qin, Zhang, Chen, Lakshminarayanan, Beutel, and Wang]{qin2022understanding}
Yao Qin, Chiyuan Zhang, Ting Chen, Balaji Lakshminarayanan, Alex Beutel, and Xuezhi Wang.
\newblock Understanding and improving robustness of vision transformers through patch-based negative augmentation.
\newblock \emph{Advances in Neural Information Processing Systems}, 35:\penalty0 16276--16289, 2022.

\bibitem[Ren et~al.(2021)Ren, Shi, Wang, and Yan]{ren2021adversarial}
Qibing Ren, Liangliang Shi, Lanjun Wang, and Junchi Yan.
\newblock Adversarial robustness via adaptive label smoothing.
\newblock 2021.

\bibitem[Rusak et~al.(2020)Rusak, Schott, Zimmermann, Bitterwolf, Bringmann, Bethge, and Brendel]{rusak2020simple}
Evgenia Rusak, Lukas Schott, Roland~S Zimmermann, Julian Bitterwolf, Oliver Bringmann, Matthias Bethge, and Wieland Brendel.
\newblock A simple way to make neural networks robust against diverse image corruptions.
\newblock In \emph{Computer Vision--ECCV 2020: 16th European Conference, Glasgow, UK, August 23--28, 2020, Proceedings, Part III 16}, pp.\  53--69. Springer, 2020.

\bibitem[Shafahi et~al.(2019)Shafahi, Ghiasi, Huang, and Goldstein]{shafahi2019label}
Ali Shafahi, Amin Ghiasi, Furong Huang, and Tom Goldstein.
\newblock Label smoothing and logit squeezing: A replacement for adversarial training?
\newblock \emph{arXiv preprint arXiv:1910.11585}, 2019.

\bibitem[Shorten \& Khoshgoftaar(2019)Shorten and Khoshgoftaar]{shorten2019survey}
Connor Shorten and Taghi~M Khoshgoftaar.
\newblock A survey on image data augmentation for deep learning.
\newblock \emph{Journal of big data}, 6\penalty0 (1):\penalty0 1--48, 2019.

\bibitem[Standley et~al.(2020)Standley, Zamir, Chen, Guibas, Malik, and Savarese]{standley2020tasks}
Trevor Standley, Amir Zamir, Dawn Chen, Leonidas Guibas, Jitendra Malik, and Silvio Savarese.
\newblock Which tasks should be learned together in multi-task learning?
\newblock In \emph{International conference on machine learning}, pp.\  9120--9132. PMLR, 2020.

\bibitem[Summers \& Dinneen(2019)Summers and Dinneen]{summers2019improved}
Cecilia Summers and Michael~J Dinneen.
\newblock Improved mixed-example data augmentation.
\newblock In \emph{2019 IEEE winter conference on applications of computer vision (WACV)}, pp.\  1262--1270. IEEE, 2019.

\bibitem[Szegedy et~al.(2013)Szegedy, Zaremba, Sutskever, Bruna, Erhan, Goodfellow, and Fergus]{szegedy2013intriguing}
Christian Szegedy, Wojciech Zaremba, Ilya Sutskever, Joan Bruna, Dumitru Erhan, Ian Goodfellow, and Rob Fergus.
\newblock Intriguing properties of neural networks.
\newblock \emph{arXiv preprint arXiv:1312.6199}, 2013.

\bibitem[Szegedy et~al.(2016)Szegedy, Vanhoucke, Ioffe, Shlens, and Wojna]{szegedy2016rethinking}
Christian Szegedy, Vincent Vanhoucke, Sergey Ioffe, Jon Shlens, and Zbigniew Wojna.
\newblock Rethinking the inception architecture for computer vision.
\newblock In \emph{Proceedings of the IEEE conference on computer vision and pattern recognition}, pp.\  2818--2826, 2016.

\bibitem[Tram{\`e}r et~al.(2017{\natexlab{a}})Tram{\`e}r, Kurakin, Papernot, Goodfellow, Boneh, and McDaniel]{tramer2017ensemble}
Florian Tram{\`e}r, Alexey Kurakin, Nicolas Papernot, Ian Goodfellow, Dan Boneh, and Patrick McDaniel.
\newblock Ensemble adversarial training: Attacks and defenses.
\newblock \emph{arXiv preprint arXiv:1705.07204}, 2017{\natexlab{a}}.

\bibitem[Tram{\`e}r et~al.(2017{\natexlab{b}})Tram{\`e}r, Papernot, Goodfellow, Boneh, and McDaniel]{tramer2017space}
Florian Tram{\`e}r, Nicolas Papernot, Ian Goodfellow, Dan Boneh, and Patrick McDaniel.
\newblock The space of transferable adversarial examples.
\newblock \emph{arXiv preprint arXiv:1704.03453}, 2017{\natexlab{b}}.

\bibitem[Tr{\"a}uble et~al.(2021)Tr{\"a}uble, Creager, Kilbertus, Locatello, Dittadi, Goyal, Sch{\"o}lkopf, and Bauer]{trauble2021disentangled}
Frederik Tr{\"a}uble, Elliot Creager, Niki Kilbertus, Francesco Locatello, Andrea Dittadi, Anirudh Goyal, Bernhard Sch{\"o}lkopf, and Stefan Bauer.
\newblock On disentangled representations learned from correlated data.
\newblock In \emph{International Conference on Machine Learning}, pp.\  10401--10412. PMLR, 2021.

\bibitem[Vasiljevic et~al.(2016)Vasiljevic, Chakrabarti, and Shakhnarovich]{vasiljevic2016examining}
Igor Vasiljevic, Ayan Chakrabarti, and Gregory Shakhnarovich.
\newblock Examining the impact of blur on recognition by convolutional networks.
\newblock \emph{arXiv preprint arXiv:1611.05760}, 2016.

\bibitem[Wang et~al.(2023{\natexlab{a}})Wang, Li, Zhao, Heng, and Zhang]{wang2023pitfall}
Deng-Bao Wang, Lanqing Li, Peilin Zhao, Pheng-Ann Heng, and Min-Ling Zhang.
\newblock On the pitfall of mixup for uncertainty calibration.
\newblock In \emph{Proceedings of the IEEE/CVF Conference on Computer Vision and Pattern Recognition}, pp.\  7609--7618, 2023{\natexlab{a}}.

\bibitem[Wang et~al.(2021)Wang, Xiao, Kossaifi, Yu, Anandkumar, and Wang]{wang2021augmax}
Haotao Wang, Chaowei Xiao, Jean Kossaifi, Zhiding Yu, Anima Anandkumar, and Zhangyang Wang.
\newblock Augmax: Adversarial composition of random augmentations for robust training.
\newblock \emph{Advances in neural information processing systems}, 34:\penalty0 237--250, 2021.

\bibitem[Wang et~al.(2023{\natexlab{b}})Wang, Pang, Du, Lin, Liu, and Yan]{wang2023better}
Zekai Wang, Tianyu Pang, Chao Du, Min Lin, Weiwei Liu, and Shuicheng Yan.
\newblock Better diffusion models further improve adversarial training.
\newblock In \emph{International Conference on Machine Learning}, pp.\  36246--36263. PMLR, 2023{\natexlab{b}}.

\bibitem[Wu et~al.(2020)Wu, Xia, and Wang]{wu2020adversarial}
Dongxian Wu, Shu-Tao Xia, and Yisen Wang.
\newblock Adversarial weight perturbation helps robust generalization.
\newblock \emph{Advances in neural information processing systems}, 33:\penalty0 2958--2969, 2020.

\bibitem[Xie et~al.(2017)Xie, Girshick, Doll{\'a}r, Tu, and He]{xie2017aggregated}
Saining Xie, Ross Girshick, Piotr Doll{\'a}r, Zhuowen Tu, and Kaiming He.
\newblock Aggregated residual transformations for deep neural networks.
\newblock In \emph{Proceedings of the IEEE conference on computer vision and pattern recognition}, pp.\  1492--1500, 2017.

\bibitem[Xin et~al.(2022)Xin, Ghorbani, Gilmer, Garg, and Firat]{xin2022current}
Derrick Xin, Behrooz Ghorbani, Justin Gilmer, Ankush Garg, and Orhan Firat.
\newblock Do current multi-task optimization methods in deep learning even help?
\newblock \emph{Advances in neural information processing systems}, 35:\penalty0 13597--13609, 2022.

\bibitem[Xu et~al.(2023)Xu, Yoon, Fuentes, and Park]{xu2023comprehensive}
Mingle Xu, Sook Yoon, Alvaro Fuentes, and Dong~Sun Park.
\newblock A comprehensive survey of image augmentation techniques for deep learning.
\newblock \emph{Pattern Recognition}, pp.\  109347, 2023.

\bibitem[Xu et~al.(2017)Xu, Evans, and Qi]{xu2017feature}
Weilin Xu, David Evans, and Yanjun Qi.
\newblock Feature squeezing: Detecting adversarial examples in deep neural networks.
\newblock \emph{arXiv preprint arXiv:1704.01155}, 2017.

\bibitem[Yao et~al.(2022)Yao, Wang, Li, Zhang, Liang, Zou, and Finn]{yao2022improving}
Huaxiu Yao, Yu~Wang, Sai Li, Linjun Zhang, Weixin Liang, James Zou, and Chelsea Finn.
\newblock Improving out-of-distribution robustness via selective augmentation.
\newblock In \emph{International Conference on Machine Learning}, pp.\  25407--25437. PMLR, 2022.

\bibitem[Zagoruyko \& Komodakis(2016)Zagoruyko and Komodakis]{zagoruyko2016wide}
Sergey Zagoruyko and Nikos Komodakis.
\newblock Wide residual networks.
\newblock \emph{arXiv preprint arXiv:1605.07146}, 2016.

\bibitem[Zhang et~al.(2017)Zhang, Cisse, Dauphin, and Lopez-Paz]{zhang2017mixup}
Hongyi Zhang, Moustapha Cisse, Yann~N Dauphin, and David Lopez-Paz.
\newblock mixup: Beyond empirical risk minimization.
\newblock \emph{arXiv preprint arXiv:1710.09412}, 2017.

\bibitem[Zhang \& Yang(2021)Zhang and Yang]{zhang2021survey}
Yu~Zhang and Qiang Yang.
\newblock A survey on multi-task learning.
\newblock \emph{IEEE Transactions on Knowledge and Data Engineering}, 34\penalty0 (12):\penalty0 5586--5609, 2021.

\bibitem[Zhong et~al.(2020)Zhong, Zheng, Kang, Li, and Yang]{zhong2020random}
Zhun Zhong, Liang Zheng, Guoliang Kang, Shaozi Li, and Yi~Yang.
\newblock Random erasing data augmentation.
\newblock In \emph{Proceedings of the AAAI conference on artificial intelligence}, volume~34, pp.\  13001--13008, 2020.

\bibitem[Zini et~al.(2022)Zini, Gomez-Villa, Buzzelli, Twardowski, Bagdanov, and van~de Weijer]{zini2022planckian}
Simone Zini, Alex Gomez-Villa, Marco Buzzelli, Bart{\l}omiej Twardowski, Andrew~D Bagdanov, and Joost van~de Weijer.
\newblock Planckian jitter: countering the color-crippling effects of color jitter on self-supervised training.
\newblock \emph{arXiv preprint arXiv:2202.07993}, 2022.

\end{thebibliography}
\bibliographystyle{collas2024_conference}

\clearpage

\appendix
\section{Appendix}
The following includes results for evaluating with AutoAttack, employing a wider range of augmentation operations when using LA, and percentages of changes for both CIFAR-10 and CIFAR-100 datasets compared to the baseline.

The following Tables~\ref{tab:percentage_cf10} and~\ref{tab:percentage_cf100} presents the percentage of improvement of methods compared to the standard training, for CIFAR-10 and CIFAR-100, respectively. Next, table~\ref{tab:MTL_more_intensity} summarizes the impact of incorporating additional operations and intensity levels.


\begin{table}[hbt!] 
\centering
\small
\renewcommand{\arraystretch}{1.0} 
\setlength{\tabcolsep}{3pt} 
\begin{tabular}{lllccccccccc}
\toprule
\multicolumn{3}{c}{\multirow{2}{*}{}} & \multicolumn{6}{c}{Train} \\
\cmidrule{4-12}
\multicolumn{3}{c}{} & \footnotesize \text{Std.} & \footnotesize \text{Mixup} & \footnotesize \text{AugMix} & \footnotesize \text{AutoAug.} & \footnotesize \text{FGSM} &  \footnotesize \text{PGD$_{10}$} & \footnotesize \text{P.G.$^{\mdoubleplus}$} & \footnotesize \text{P.G.J.$^{\mdoubleplus}$} & \footnotesize \text{AugMix$^{\mdoubleplus}$}\\
\midrule
  &
\multirow{4}{*}{\footnotesize {Clean}} 
& \footnotesize LeNet & 0.00 & 12.29 & -1.00 & -3.21 & -0.94 & 77.49 & -5.74 & -13.89 & 17.90\\
&& \footnotesize ResNet & 0.00 & 3.68 & -1.89 & -9.34 & 5.27 & 117.10 & -11.63 & -15.90 & -5.27\\
&& \footnotesize ResNeXt & 0.00 & 3.52 & -3.52 & -16.80 & 3.96 & 115.83 & -19.26 & -24.01 & -5.54\\
&& \footnotesize WResNet & 0.00 & 8.67 & -1.67 & -12.85 & 7.31 & 162.59 & -16.82 & -15.57 & -0.94\\
&& \footnotesize SwinT & 0.00 & -15.10 & -28.72 & -5.69 & -28.44 & 14.61 & -41.64 & -42.98 & -28.72\\
\arrayrulecolor{black!20}\cmidrule{3-12}
&& \footnotesize Mean & 0.00 & 2.14 & -8.29 & -8.92 & -4.17 & 90.00 & -19.53 & -23.29 & -4.42\\
\arrayrulecolor{black}\cmidrule{2-12}
&
\multirow{4}{*}{\footnotesize {mCE}} 
& \footnotesize LeNet & 0.00 & -0.11 & -15.58 & -10.94 & -0.74 & 43.98 & -14.84 & -16.04 & 9.35\\
&& \footnotesize ResNet & 0.00 & -7.12 & -24.28 & -21.21 & -9.35 & 51.72 & -24.89 & -25.05 & -4.05\\
&& \footnotesize ResNeXt & 0.00 & -3.91 & -24.49 & -23.00 & -10.27 & 63.11 & -25.45 & -27.10 & 6.52\\
&& \footnotesize WResNet & 0.00 & -5.19 & -25.16 & -21.29 & -7.66 & 87.33 & -26.79 & -26.18 & 0.09\\
&& \footnotesize SwinT & 0.00 & 0.58 & -22.43 & -0.13 & -12.59 & 1.17 & -28.30 & -28.22 & -22.43\\
\arrayrulecolor{black!20}\cmidrule{3-12}
&& \footnotesize Mean & 0.00 & -3.05 & -22.15 & -15.19 & -7.90 & 49.00 & -23.73 & -24.23 & -1.69\\
\arrayrulecolor{black}\cmidrule{2-12}
\multirow{12}{*}{\rotatebox{90}{\parbox{1.1cm}{\small Calibration}}} 
&\multirow{4}{*}{\small {ECE}} 
& \footnotesize LeNet & 0.00 & 125.77 & -24.15 & -57.00 & -4.86 & 125.18 & -8.98 & 12.67 & -67.89\\
&& \footnotesize ResNet & 0.00 & 169.48 & -14.47 & -41.34 & -4.77 & 159.46 & -7.47 & -7.63 & -51.67\\
&& \footnotesize ResNeXt & 0.00 & 148.18 & -20.83 & -48.44 & 0.52 & 150.65 & -19.53 & -24.09 & -28.65\\
&& \footnotesize WResNet & 0.00 & 192.39 & -12.04 & -34.55 & 2.22 & 298.73 & -19.02 & -1.90 & -2.22\\
&& \footnotesize SwinT & 0.00 & 111.38 & -19.85 & -35.35 & 18.89 & -26.15 & 25.18 & 22.03 & -7.26\\
\arrayrulecolor{black!20}\cmidrule{3-12}
&& \footnotesize Mean & 0.00 & 151.67 & -18.37 & -44.33 & 1.06 & 153.43 & -8.78 & -2.18 & -33.65\\
\arrayrulecolor{black}\cmidrule{3-12}
&\multirow{4}{*}{\small {RMS}} 
& \footnotesize LeNet & 0.00 & 65.98 & -21.06 & -50.62 & -7.47 & 87.66 & -3.42 & -1.56 & -61.93\\
&& \footnotesize ResNet & 0.00 & 69.31 & -14.13 & -36.40 & -7.45 & 93.32 & -3.97 & -11.13 & -51.31\\
&& \footnotesize ResNeXt & 0.00 & 64.18 & -15.34 & -41.63 & -4.56 & 88.31 & -16.92 & -20.73 & -30.76\\
&& \footnotesize WResNet & 0.00 & 85.25 & -11.09 & -25.65 & -1.35 & 175.41 & -17.45 & -13.60 & -4.44\\
&& \footnotesize SwinT & 0.00 & 49.07 & 0.62 & -27.31 & -3.70 & 36.57 & 39.04 & 22.69 & 0.62\\
\arrayrulecolor{black!20}\cmidrule{3-12}
&& \footnotesize Mean & 0.00 & 68.09 & -13.20 & -37.01 & -4.95 & 100.86 & -4.21 & -7.65 & -31.51\\
\arrayrulecolor{black}\cmidrule{2-12}
\multirow{12}{*}{\rotatebox{90}{\parbox{3cm}{\small Adversarial (\scriptsize$\varepsilon=0.03)$}}} &
\multirow{4}{*}{ {\small FGSM}} 
& \footnotesize LeNet & 0.00 & -9.11 & -2.86 & -0.44 & -20.50 & -26.51 & -46.08 & -53.94 & -12.20\\
&& \footnotesize ResNet & 0.00 & -23.12 & 0.65 & 8.33 & -16.06 & 7.15 & -55.49 & -58.76 & -22.58\\
&& \footnotesize ResNeXt & 0.00 & -17.08 & 0.26 & 8.36 & -20.47 & 11.82 & -54.77 & -57.94 & -22.41\\
&& \footnotesize WResNet & 0.00 & -25.75 & 0.86 & 1.67 & -19.75 & 15.02 & -57.46 & -66.53 & -31.36\\
&& \footnotesize SwinT & 0.00 & 7.56 & 1.49 & 12.79 & -6.74 & -65.26 & -65.61 & -68.05 & -54.62\\
\arrayrulecolor{black!20}\cmidrule{3-12}
&& \footnotesize Mean & 0.00 & -11.91 & 0.07 & 6.36 & -16.15 & -15.83 & -56.17 & -61.24 & -29.78\\
\arrayrulecolor{black}\cmidrule{3-12}
&\multirow{4}{*}{\small {PGD$_{40}$}} 
& \footnotesize LeNet & 0.00 & -4.22 & -2.71 & -1.91 & -18.35 & -29.33 & -44.45 & -49.80 & -21.60\\
&& \footnotesize ResNet & 0.00 & -19.09 & 0.69 & 4.33 & -9.26 & -10.17 & -46.39 & -51.76 & -22.47\\
&& \footnotesize ResNeXt & 0.00 & -13.26 & -0.70 & 6.73 & -13.14 & -9.47 & -51.02 & -53.61 & -30.09\\
&& \footnotesize WResNet & 0.00 & -18.75 & 1.05 & 4.42 & -11.73 & -7.86 & -48.39 & -50.95 & -37.26\\
&& \footnotesize SwinT & 0.00 & 20.01 & -7.03 & 14.26 & -1.75 & -70.21 & -31.65 & -41.12 & -7.03\\
\arrayrulecolor{black!20}\cmidrule{3-12}
&& \footnotesize Mean & 0.00 & -7.40 & -1.67 & 5.49 & -10.91 & -24.88 & -44.52 & -49.54 & -23.87\\
\arrayrulecolor{black}\cmidrule{2-12}
\multirow{12}{*}{\rotatebox{90}{\parbox{3cm}{\small Adversarial (\scriptsize$\varepsilon=0.3)$}}} &
\multirow{4}{*}{ {\small FGSM}} 
& \footnotesize LeNet & 0.00 & -9.24 & -0.51 & 3.68 & -38.81 & -40.63 & -41.98 & -45.36 & -39.14\\
&& \footnotesize ResNet & 0.00 & -17.84 & -0.50 & 8.64 & -34.41 & -56.77 & -45.75 & -51.88 & -31.42\\
&& \footnotesize ResNeXt & 0.00 & -16.76 & 1.16 & 7.22 & -54.95 & -56.76 & -48.64 & -53.18 & -32.14\\
&& \footnotesize WResNet & 0.00 & -17.09 & 5.42 & 12.25 & -44.30 & -60.36 & -44.51 & -50.29 & -34.49\\
&& \footnotesize SwinT & 0.00 & -4.11 & -0.15 & 1.56 & -61.68 & -12.47 & -58.80 & -63.88 & -8.23\\
\arrayrulecolor{black!20}\cmidrule{3-12}
&& \footnotesize Mean & 0.00 & -12.50 & 0.93 & 6.31 & -47.22 & -43.77 & -48.25 & -53.18 & -28.47\\
\arrayrulecolor{black}\cmidrule{3-12}
&\multirow{4}{*}{\small {PGD$_{40}$}} 
& \footnotesize LeNet & 0.00 & -1.30 & 3.51 & 0.94 & 0.10 & -12.27 & -26.33 & -28.89 & -9.29\\
&& \footnotesize ResNet & 0.00 & -12.70 & -0.08 & 0.75 & -0.33 & -14.37 & -25.54 & -30.24 & -18.84\\
&& \footnotesize ResNeXt & 0.00 & -8.47 & -0.39 & 1.17 & -0.82 & -16.51 & -30.61 & -32.99 & -23.83\\
&& \footnotesize WResNet & 0.00 & -12.07 & 0.01 & 0.79 & -0.78 & -18.39 & -26.99 & -28.27 & -24.91\\
&& \footnotesize SwinT & 0.00 & 2.24 & 2.53 & 2.16 & 2.27 & -3.97 & 1.83 & -1.59 & -1.94\\
\arrayrulecolor{black!20}\cmidrule{3-12}
&& \footnotesize Mean & 0.00 & -6.52 & 1.10 & 1.16 & 0.08 & -13.14 & -21.59 & -24.46 & -15.85\\
\arrayrulecolor{black}\bottomrule
\end{tabular}
\caption{Percentages of Error rates of various methods across different architectures for CIFAR-10 when compared to Standard training. Negative values indicate an improvement in error rates when employing augmentation techniques.}
\label{tab:percentage_cf10}
\end{table}

\clearpage


\begin{table}[hbt!] 
\centering
\small
\renewcommand{\arraystretch}{1.0} 
\setlength{\tabcolsep}{3pt} 
\begin{tabular}{lllccccccccc}
\toprule
\multicolumn{3}{c}{\multirow{2}{*}{}} & \multicolumn{6}{c}{Train} \\
\cmidrule{4-12}
\multicolumn{3}{c}{} & \footnotesize \text{Std.} & \footnotesize \text{Mixup} & \footnotesize \text{AugMix} & \footnotesize \text{AutoAug.} & \footnotesize \text{FGSM} &  \footnotesize \text{PGD$_{10}$} & \footnotesize \text{P.G.$^{\mdoubleplus}$} & \footnotesize \text{P.G.J.$^{\mdoubleplus}$} & \footnotesize \text{AugMix$^{\mdoubleplus}$}\\
\midrule
  &
\multirow{4}{*}{\footnotesize {Clean}} 
& \footnotesize LeNet & 0.00 & 4.29 & -1.08 & -2.38 & 5.25 & 43.05 & -2.61 & -1.56 & 0.58\\
&& \footnotesize ResNet & 0.00 & 3.67 & -0.74 & -3.83 & 9.23 & 52.64 & -1.38 & -3.05 & -4.44\\
&& \footnotesize ResNeXt & 0.00 & 5.89 & -2.79 & -9.89 & 8.22 & 54.84 & -6.86 & -8.92 & -6.16\\
&& \footnotesize WResNet & 0.00 & 6.63 & -1.77 & -7.17 & 9.62 & 44.10 & -2.75 & -2.85 & -6.73\\
&& \footnotesize SwinT & 0.00 & 4.05 & -1.61 & -1.61 & 10.78 & 25.14 & -2.47 & -5.73 & -1.90\\
\arrayrulecolor{black!20}\cmidrule{3-12}
&& \footnotesize Mean & 0.00 & 4.88 & -1.58 & -4.94 & 8.38 & 44.40 & -3.24 & -4.31 & -3.53\\
\arrayrulecolor{black}\cmidrule{2-12}
&
\multirow{4}{*}{\footnotesize {mCE}} 
& \footnotesize LeNet & 0.00 & -1.43 & -9.25 & -7.96 & -5.40 & 15.98 & -6.39 & -6.74 & -4.03\\
&& \footnotesize ResNet & 0.00 & -4.51 & -15.81 & -13.12 & -8.16 & 10.70 & -11.80 & -12.29 & -10.85\\
&& \footnotesize ResNeXt & 0.00 & -2.36 & -14.25 & -14.52 & -6.10 & 21.95 & -11.33 & -12.77 & -11.73\\
&& \footnotesize WResNet & 0.00 & 0.99 & -13.55 & -11.48 & -3.29 & 18.04 & -7.97 & -8.32 & -10.18\\
&& \footnotesize SwinT & 0.00 & -1.42 & -10.63 & -7.17 & -4.33 & -1.25 & -2.19 & -9.53 & -5.42\\
\arrayrulecolor{black!20}\cmidrule{3-12}
&& \footnotesize Mean & 0.00 & -1.80 & -12.68 & -10.89 & -5.52 & 13.44 & -8.06 & -9.91 & -8.42\\
\arrayrulecolor{black}\cmidrule{2-12}
\multirow{12}{*}{\rotatebox{90}{\parbox{1.1cm}{\small Calibration}}} 
&\multirow{4}{*}{\small {ECE}} 
& \footnotesize LeNet & 0.00 & -35.40 & -27.23 & -59.66 & -1.39 & 60.35 & -10.83 & -14.69 & -32.81\\
&& \footnotesize ResNet & 0.00 & -47.88 & -14.35 & -33.04 & -15.38 & 59.35 & -15.71 & -23.10 & -30.82\\
&& \footnotesize ResNeXt & 0.00 & -45.51 & -15.89 & -38.50 & 3.89 & 61.79 & -25.64 & -31.64 & -38.21\\
&& \footnotesize WResNet & 0.00 & -45.43 & -10.64 & -29.34 & 6.75 & 48.68 & -8.32 & -11.35 & -35.82\\
&& \footnotesize SwinT & 0.00 & -55.02 & -19.45 & -48.30 & 3.54 & 36.99 & -0.85 & -14.29 & -21.57\\
\arrayrulecolor{black!20}\cmidrule{3-12}
&& \footnotesize Mean & 0.00 & -45.71 & -17.08 & -40.81 & -0.56 & 54.25 & -13.23 & -19.71 & -32.50\\
\arrayrulecolor{black}\cmidrule{3-12}
&\multirow{4}{*}{\small {RMS}} 
& \footnotesize LeNet & 0.00 & -38.42 & -26.26 & 52.36 & -1.90 & 49.70 & -13.13 & -5.37 & -30.49\\
&& \footnotesize ResNet & 0.00 & -50.75 & -13.12 & -30.30 & 0.23 & 46.00 & -34.28 & -18.41 & -29.17\\
&& \footnotesize ResNeXt & 0.00 & -49.07 & -14.16 & -35.59 & 3.04 & 47.69 & -36.53 & -36.97 & -36.24\\
&& \footnotesize WResNet & 0.00 & -51.13 & -10.99 & -29.03 & 4.23 & 34.26 & -33.83 & -20.88 & -32.91\\
&& \footnotesize SwinT & 0.00 & -57.00 & -17.01 & -46.59 & 1.59 & 30.03 & -28.78 & -17.58 & -13.42\\
\arrayrulecolor{black!20}\cmidrule{3-12}
&& \footnotesize Mean & 0.00 & -49.33 & -15.84 & -19.56 & 1.61 & 41.84 & -30.10 & -20.86 & -29.24\\
\arrayrulecolor{black}\cmidrule{2-12}
\multirow{12}{*}{\rotatebox{90}{\parbox{3cm}{\small Adversarial (\scriptsize$\varepsilon=0.03)$}}} &
\multirow{4}{*}{ {\small FGSM}} 
& \footnotesize LeNet & 0.00 & -2.74 & -1.01 & 1.74 & -8.83 & -7.64 & -17.95 & -17.65 & -6.38\\
&& \footnotesize ResNet & 0.00 & -9.49 & -1.76 & 2.15 & -10.19 & 1.20 & -27.09 & -27.82 & -19.33\\
&& \footnotesize ResNeXt & 0.00 & -6.20 & 1.37 & 1.95 & -9.29 & 3.15 & -25.16 & -26.39 & -19.79\\
&& \footnotesize WResNet & 0.00 & -10.63 & 0.31 & 2.50 & -9.73 & 0.52 & -22.53 & -29.87 & -22.11\\
&& \footnotesize SwinT & 0.00 & -0.76 & -0.16 & 4.10 & -11.13 & -34.45 & -37.47 & -32.79 & -34.72\\
\arrayrulecolor{black!20}\cmidrule{3-12}
&& \footnotesize Mean & 0.00 & -5.71 & -0.27 & 2.53 & -9.87 & -8.46 & -26.32 & -26.94 & -20.71\\
\arrayrulecolor{black}\cmidrule{3-12}
&\multirow{4}{*}{\small {PGD$_{40}$}} 
& \footnotesize LeNet & 0.00 & -0.56 & -0.54 & 1.87 & -6.71 & -11.54 & -13.01 & -11.83 & -3.49\\
&& \footnotesize ResNet & 0.00 & -4.56 & 0.31 & 3.26 & -7.29 & -5.99 & -22.98 & -23.62 & -11.93\\
&& \footnotesize ResNeXt & 0.00 & -3.83 & 1.36 & 3.95 & -6.97 & -5.01 & -19.74 & -19.33 & -13.89\\
&& \footnotesize WResNet & 0.00 & -6.89 & 0.22 & 2.27 & -6.71 & -7.69 & -23.06 & -25.11 & -13.66\\
&& \footnotesize SwinT & 0.00 & 5.60 & 1.20 & 4.70 & -11.91 & -37.22 & -25.02 & -12.43 & -15.74\\
\arrayrulecolor{black!20}\cmidrule{3-12}
&& \footnotesize Mean & 0.00 & -1.99 & 0.51 & 3.21 & -7.94 & -13.66 & -20.77 & -18.41 & -11.74\\
\arrayrulecolor{black}\cmidrule{2-12}
\multirow{12}{*}{\rotatebox{90}{\parbox{3cm}{\small Adversarial (\scriptsize$\varepsilon=0.3)$}}} &
\multirow{4}{*}{ {\small FGSM}} 
& \footnotesize LeNet & 0.00 & -1.29 & -2.12 & 1.08 & -20.00 & -24.40 & -10.96 & -10.96 & -4.83\\
&& \footnotesize ResNet & 0.00 & -3.72 & -3.01 & -0.34 & -21.18 & -28.30 & -14.56 & -16.58 & -13.00\\
&& \footnotesize ResNeXt & 0.00 & -2.89 & -4.91 & -1.72 & -20.10 & -27.43 & -14.36 & -18.68 & -10.50\\
&& \footnotesize WResNet & 0.00 & -1.78 & -3.51 & -0.96 & -24.00 & -28.96 & -16.19 & -20.59 & -13.09\\
&& \footnotesize SwinT & 0.00 & -1.65 & -1.41 & 1.26 & -55.32 & -13.46 & -19.12 & -18.20 & -18.40\\
\arrayrulecolor{black!20}\cmidrule{3-12}
&& \footnotesize Mean & 0.00 & -2.26 & -2.97 & -0.11 & -28.29 & -24.40 & -15.03 & 7.00 & -11.95\\
\arrayrulecolor{black}\cmidrule{3-12}
&\multirow{4}{*}{\small {PGD$_{40}$}} 
& \footnotesize LeNet & 0.00 & 1.77 & -0.10 & 1.84 & 0.20 & -9.97 & -6.45 & -6.92 & -0.58\\
&& \footnotesize ResNet & 0.00 & -1.72 & -1.37 & 0.43 & -0.57 & -12.80 & -12.69 & -13.40 & -4.34\\
&& \footnotesize ResNeXt & 0.00 & -0.28 & -1.09 & 1.06 & -0.08 & -14.01 & -9.34 & -9.41 & -6.93\\
&& \footnotesize WResNet & 0.00 & -2.75 & -0.97 & 0.64 & -0.84 & -11.02 & -10.84 & -4.77 & -5.71\\
&& \footnotesize SwinT & 0.00 & 1.74 & -0.56 & 1.82 & -1.80 & -2.43 & -12.53 & -1.12 & -5.95\\
\arrayrulecolor{black!20}\cmidrule{3-12}
&& \footnotesize Mean & 0.00 & -0.25 & -0.82 & 1.16 & -0.62 & -10.03 & -10.38 & -7.11 & -4.71\\
\arrayrulecolor{black}\bottomrule
\end{tabular}
\caption{Percentages of Error rates of various methods across different architectures for CIFAR-100 when compared to Standard training. Negative values indicate an improvement in error rates when employing augmentation techniques.}
\label{tab:percentage_cf100}
\end{table}

\clearpage

\textbf{The impact of incorporating additional operations and intensity levels.} 

Table~\ref{tab:MTL_more_intensity} presents the results obtained by employing additional augmentation operations and intensity levels while utilizing LA.

\begin{table}[h]
\centering
\small 
\renewcommand{\arraystretch}{1.3} 
\setlength{\tabcolsep}{5pt} 
\begin{minipage}[t]{0.45\linewidth}
\centering
\begin{tabular}{@{}clcc@{}}
\toprule
\small & & \multicolumn{2}{c}{\small Train} \\ \cmidrule(lr){3-4}
\small & Err. & \small Std.  & \small LA. \\
\midrule
& Clean 
&\footnotesize 9.57\tiny\textcolor{gray!95}{$\pm$0.21} & 6.79\tiny\textcolor{gray!95}{$\pm$1.72} \\
& mCE 
&\footnotesize 22.73\tiny\textcolor{gray!95}{$\pm$0.27} & 13.72\tiny\textcolor{gray!95}{$\pm$1.07} \\
\arrayrulecolor{black!20}\cmidrule{2-4}
& ECE  
&\footnotesize 6.31\tiny\textcolor{gray!95}{$\pm$0.21} & 5.23\tiny\textcolor{gray!95}{$\pm$0.66} \\
& RMS 
&\footnotesize 10.37\tiny\textcolor{gray!95}{$\pm$0.38} & 9.26\tiny\textcolor{gray!95}{$\pm$0.40} \\
\arrayrulecolor{black!20}\cmidrule{2-4}
\multirow{2}{*}{\rotatebox{90}{\small \parbox{1.1cm}{\centering \tiny$\varepsilon \tiny= \tiny0.03$}}}
& FGSM 
&\footnotesize 43.14\tiny\textcolor{gray!95}{$\pm$0.23} & 14.15\tiny\textcolor{gray!95}{$\pm$4.24} \\
& PGD 
&\footnotesize 77.08\tiny\textcolor{gray!95}{$\pm$0.18} & 35.04\tiny\textcolor{gray!95}{$\pm$0.44} \\
\arrayrulecolor{black!20}\cmidrule{2-4}
\multirow{2}{*}{\rotatebox{90}{\small \parbox{1.1cm}{\centering \tiny$\varepsilon \tiny= \tiny0.3$}}}
& FGSM 
&\footnotesize 69.53\tiny\textcolor{gray!95}{$\pm$0.88} & 25.89\tiny\textcolor{gray!95}{$\pm$5.69} \\
& PGD 
&\footnotesize 94.78\tiny\textcolor{gray!95}{$\pm$0.53} & 50\tiny\textcolor{gray!95}{$\pm$3.20} \\ 
\arrayrulecolor{black!20}\cmidrule{2-4}
& AA
&\footnotesize 86.05\tiny\textcolor{gray!95}{$\pm$-} & 72.07\tiny\textcolor{gray!95}{$\pm$-} \\ 
\arrayrulecolor{black}\cmidrule{2-4}
\end{tabular}
\caption{\textbf{CIFAR-10}}
\label{tab:CIFAR-10}
\end{minipage}
\begin{minipage}[t]{0.45\linewidth}
\centering
\setlength{\tabcolsep}{5pt} 
\begin{tabular}{@{}clcc@{}}
\toprule
\small & & \multicolumn{2}{c}{\small Train} \\ \cmidrule(lr){3-4}
\small & Err. & \small Std.  & \small LA. \\
\midrule
& Clean 
&\footnotesize 29.43\tiny\textcolor{gray!95}{$\pm$0.19} & 27.74\tiny\textcolor{gray!95}{$\pm$1.26} \\
& mCE 
&\footnotesize 48.33\tiny\textcolor{gray!95}{$\pm$0.25} & 40.52\tiny\textcolor{gray!95}{$\pm$1.67} \\
\arrayrulecolor{black!20}\cmidrule{2-4}
& ECE  
&\footnotesize 18.51\tiny\textcolor{gray!95}{$\pm$0.33} & 18.13\tiny\textcolor{gray!95}{$\pm$0.77} \\
& RMS 
&\footnotesize 22.94\tiny\textcolor{gray!95}{$\pm$0.46} & 22.37\tiny\textcolor{gray!95}{$\pm$0.59} \\
\arrayrulecolor{black!20}\cmidrule{2-4}
\multirow{2}{*}{\rotatebox{90}{\small \parbox{1.1cm}{\centering \tiny$\varepsilon \tiny= \tiny0.03$}}}
& FGSM 
&\footnotesize 61.97\tiny\textcolor{gray!95}{$\pm$0.42} & 39.65\tiny\textcolor{gray!95}{$\pm$3.25} \\
& PGD 
&\footnotesize 82.01\tiny\textcolor{gray!95}{$\pm$0.35} & 56.82\tiny\textcolor{gray!95}{$\pm$1.04} \\
\arrayrulecolor{black!20}\cmidrule{2-4}
\multirow{2}{*}{\rotatebox{90}{\small \parbox{1.1cm}{\centering \tiny$\varepsilon \tiny= \tiny0.3$}}}
& FGSM 
&\footnotesize 85.46\tiny\textcolor{gray!95}{$\pm$0.82} & 56.67\tiny\textcolor{gray!95}{$\pm$2.88} \\
& PGD 
&\footnotesize 93.11\tiny\textcolor{gray!95}{$\pm$0.79} & 71.36\tiny\textcolor{gray!95}{$\pm$2.75} \\
\arrayrulecolor{black!20}\cmidrule{2-4}
& AA
&\footnotesize 93.97\tiny\textcolor{gray!95}{$\pm$-} & 82.22\tiny\textcolor{gray!95}{$\pm$-} \\
\arrayrulecolor{black}\cmidrule{2-4}
\end{tabular}
\caption{\textbf{CIFAR-100}}
\label{tab:CIFAR-100}
\end{minipage}
\caption{The error rates of Wide ResNet-50 architectures when employing LA with seven augmentation types: Plasma, Gamma, PlanckianJitter, ColorJiggle, Equalize, Posterize, and Rain across three severity levels. The parameter $\delta$ corresponds to the intensity of the added noise, setting $\delta=0.8,0.6,0.2$, for high, moderate, and slight noise. Similar trends in error rate improvements are observed with additional augmentation types. While improving the baseline error, in some cases, it exhibit a lower degree of improvement compared to employing fewer numbers of operations.}
\label{tab:MTL_more_intensity}
\end{table}

\clearpage



\textbf{The performance under AutoAttack.} 

AutoAttack~\citep{croce2020reliable} ensembles two parameter-free versions of the PGD attack, along with the Fast Adaptive Boundary attack~\citep{croce2020minimally} and the Square Attack~\citep{andriushchenko2020square}, to create a diverse testing framework.
We evaluate using AutoAttack with $L_\infty$ constraints at $\varepsilon = 0.03$. We use the implementations provided by the \textit{TorchAttack} 3.5.1 library~\citep{kim2020torchattacks}.

Tables~\ref{tab:cf10_AA},~\ref{tab:CIFAR-10_MTLAA}, and~\ref{tab:CIFAR-100_MTL_AA} present a summary of the performance of various training methods under AutoAttack across both the CIFAR-10 and CIFAR-100 datasets. While the robustness of the methods falls below their performance under FGSM or PGD attacks, LA consistently outperforms all other methods. Specifically, in comparison to standard training error, Mixup and PGD worsen it by 1.02\% and 6.29\%, respectively, while Augmix, AutoAugment, FGSM, PGD, P.G.$^{\mdoubleplus}$, P.G.J.$^{\mdoubleplus}$, and AugMix$^{\mdoubleplus}$ improve it by 2.02\%, 1.66\%, 4.44\%, 7.61\%, 11.76\%, and 13.76\%, respectively, for CIFAR-10. For CIFAR-100, PGD shows a slight increase of 0.19\%, whereas all other methods improve it by 0.73\%, 1.63\%, 1.78\%, 1.49\%, 6.68\%, 5.64\%, and 11.93\% for Mixup, Augmix, AutoAugment, FGSM, P.G.$^{\mdoubleplus}$, P.G.J.$^{\mdoubleplus}$, and AugMix$^{\mdoubleplus}$, respectively. In LS and MTL, the improvement under AA are 2.12\% and 4.37\% for CIFAR-10 and 1.11\% and 1.49\% for CIFAR-100, respectively.

\begin{table}[hbt!] 
\centering
\renewcommand{\arraystretch}{1.5} 
\setlength{\tabcolsep}{3pt} 
\begin{tabular}{lllccccccccc}
\toprule
\multicolumn{3}{c}{\multirow{2}{*}{}} & \multicolumn{6}{c}{\footnotesize Train} \\
\cmidrule{4-12}
\multicolumn{3}{c}{} & \footnotesize Std. & \footnotesize Mixup & \footnotesize AugMix & \footnotesize AutoAug. & \footnotesize FGSM &  \footnotesize PGD$_{10}$ & \footnotesize P.G.$^{\mdoubleplus}$ & \footnotesize P.G.J.$^{\mdoubleplus}$ & \footnotesize AugMix$^{\mdoubleplus}$ \\
\midrule
& CIFAR-10 & & 86.05 & 86.93 & 84.31 & 84.62 & 82.23 & 91.46 & 79.50 & 75.93 & 74.21 \\
& CIFAR-100 & &93.97 &93.28 &92.44 &92.30 &92.57 &94.15 &87.69 &88.67 &82.76 \\
\arrayrulecolor{black}\bottomrule
\end{tabular}
\vspace{1pt}
\caption{The error rates of different methods using Wide ResNet-50 architectures for the CIFAR-10 dataset under AutoAttack with $L_\infty$ constraints at $\varepsilon = 0.03$.}
\label{tab:cf10_AA}
\end{table}


\begin{table}[h]
\centering
\small 
\renewcommand{\arraystretch}{1.3} 
\setlength{\tabcolsep}{3pt} 
\begin{minipage}[t]{0.45\linewidth} 
\centering
\begin{tabular}{@{}clcccc@{}}
\toprule
\small & & \multicolumn{3}{c}{\small Train} \\ \cmidrule(lr){3-6}
\small & Err. & \small Std.  & \small LS. & \small MTL. & \small \text{P.G.J.$^{\mdoubleplus}$}\\
\midrule
& AA
&\footnotesize86.05 
&\footnotesize84.23 
&\footnotesize82.29
&\footnotesize75.93\\
\arrayrulecolor{black}\cmidrule{2-6}
\end{tabular}
\caption{\textbf{CIFAR-10} The error rates of MTL, LS, and LA methods using Wide ResNet-50 architectures for the CIFAR-10 dataset under AutoAttack with $L_\infty$ constraints at $\varepsilon = 0.03$.}
\label{tab:CIFAR-10_MTLAA}
\end{minipage}
\hspace{0.05\linewidth} 
\begin{minipage}[t]{0.45\linewidth} 
\centering
\setlength{\tabcolsep}{3pt} 
\begin{tabular}{@{}clcccc@{}}
\toprule
\small & & \multicolumn{3}{c}{\small Train} \\ 
\cmidrule(lr){3-6}
\small & Err. & \small Std.  & \small LS. & \small MTL. & \small \text{P.G.J.$^{\mdoubleplus}$}\\
\midrule
& AA
&\footnotesize93.97 
&\footnotesize92.93
&\footnotesize92.57
&\footnotesize88.67\\
\arrayrulecolor{black}\cmidrule{2-6}
\end{tabular}
\caption{\textbf{CIFAR-100} The error rates of MTL, LS, and LA methods using Wide ResNet-50 architectures for the CIFAR-100 dataset under AutoAttack with $L_\infty$ constraints at $\varepsilon = 0.03$.}
\label{tab:CIFAR-100_MTL_AA}
\end{minipage}
\end{table}

\clearpage

\textbf{Percentages of error rate variations compared to the standard training.}

\begin{figure}[h]
\begin{center}
\includegraphics[scale=0.35]{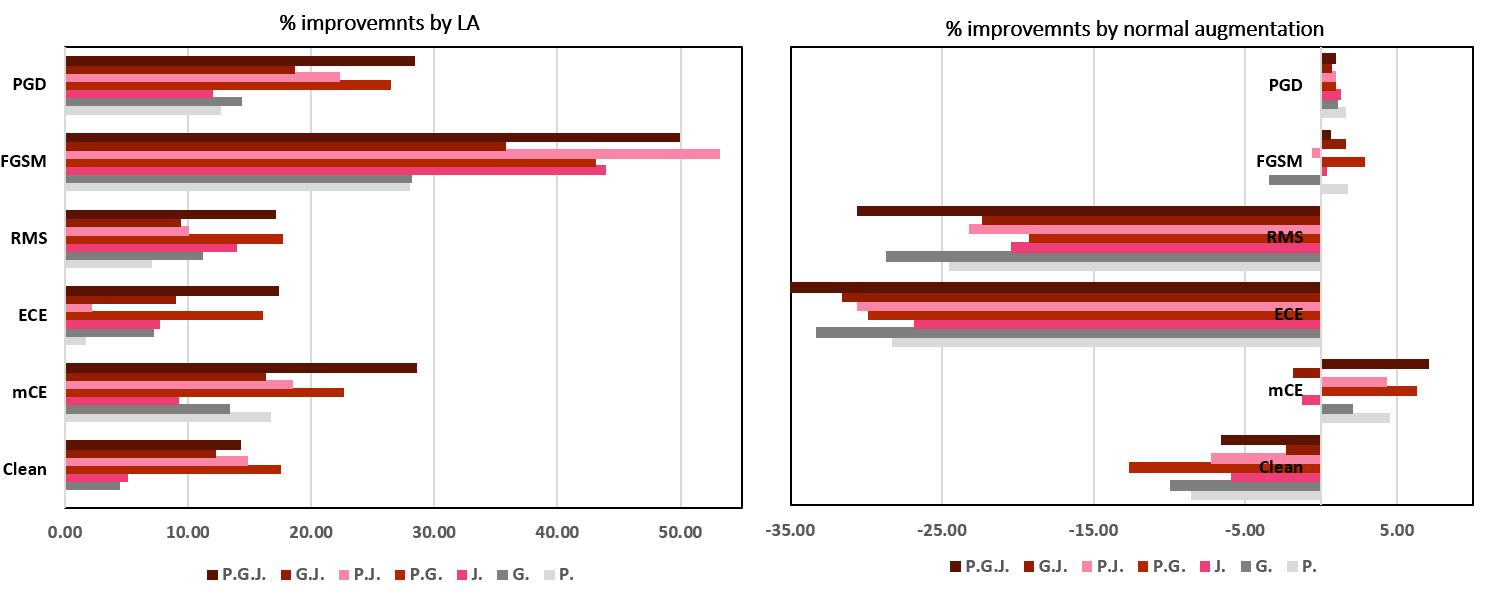}
\label{fig:WRN_LA_vs_AUG}
\end{center}
\caption{ Percentages of error rate variations compared to the standard training on Wide ResNet-50: left side employing LA, right side with normal augmentations. While normal augmentation can enhance mCE to a considerable degree, it comes at the expense of Clean and calibration errors. On the other hand, regardless of the type and the number of operations used in augmenting with LA, we can see improvements in Clean, mCE, calibration, and adversarial errors. However, using two or three types of operations proves even more effective.} 
\end{figure}

\end{document}